\documentclass[pdflatex,sn-mathphys-num]{sn-jnl}
\usepackage{graphicx}
\usepackage{amsmath, amssymb}
\usepackage{algorithm}
\usepackage{algpseudocode}
\usepackage{placeins}
\usepackage{booktabs}
\DeclareMathOperator*{\argmin}{arg\,min}
\DeclareMathOperator{\Quantile}{Quantile}
\DeclareMathOperator{\BinIndex}{BinIndex}

\usepackage[nameinlink,noabbrev,capitalise]{cleveref}

\hypersetup{
colorlinks=true,
linkcolor=blue,
citecolor=blue,
urlcolor=blue
}

\usepackage[T1]{fontenc}
\usepackage{microtype}
\usepackage[none]{hyphenat}
\title{Expectation and Acoustic Neural Network Representations Enhance Music Identification from Brain Activity}

\author{
Shogo Noguchi$^{1,*,a}$,
Taketo Akama$^{1,*}$,
Tai Nakamura$^{1,a}$, \\
Shun Minamikawa$^{1}$,
and Natalia Polouliakh$^{1}$
}

\begin{document}

\raggedbottom
\hyphenpenalty=10000
\exhyphenpenalty=10000
\doublehyphendemerits=1000000
\finalhyphendemerits=1000000

\maketitle

\begin{center}
$^{1}$Sony Computer Science Laboratories, Inc., Tokyo, Japan

\vspace{0.5em}

$^{*}$Equal contribution, corresponding authors: noguchishogo1@gmail.com, taketo.akama@gmail.com\\
$^{a}$Work conducted when working as a research assistant

\vspace{0.5em}

\end{center}

\begin{abstract}

During music listening, cortical activity encodes both acoustic and expectation-related information. 
Prior work has shown that ANN representations resemble cortical representations and can serve as supervisory signals for EEG recognition. 
Here we show that distinguishing acoustic and expectation-related ANN representations as teacher targets improves EEG-based music identification. 
Models pretrained to predict either representation outperform non-pretrained baselines, and combining them yields complementary gains that exceed strong seed ensembles formed by varying random initializations. 
These findings show that teacher representation type shapes downstream performance and that representation learning can be guided by neural encoding.
This work points toward advances in predictive music cognition and neural decoding.
Our expectation representation, computed directly from raw signals without manual labels, reflects predictive structure beyond onset or pitch, enabling investigation of multilayer predictive encoding across diverse stimuli.
Its scalability to large, diverse datasets further suggests potential for developing general-purpose EEG models grounded in cortical encoding principles.
\end{abstract}
\section{Introduction}

Prediction and expectation have long been central concepts in music cognition and neuroscience for explaining how listeners understand music as it unfolds over time. In predictive-coding accounts and related free-energy formulations, perception is treated as an active inferential process in which sensory input is continuously compared with internally generated predictions, and music has often been regarded as a particularly effective domain for studying such predictive processing \cite{Friston2010,Friston2013,Koelsch2019,Vuust2022}.
Vuust et al.\ further describe musical expectations as being evoked by auditory bottom-up sensations while also depending on the brain's top-down predictions \cite{Vuust2022}.
In the same study, predictions are described as being generated from expected states of the world and compared with observed input, which further clarifies the relation between expectation and prediction \cite{Vuust2022}.
 
Research on tonality, melody, and musical syntax likewise suggests that listeners acquire such regularities through enculturation and implicit/statistical learning, thereby forming expectations about future musical continuations \cite{Patel2003,Rohrmeier2011,Tillmann2014}. Through this learning of regularities, listeners can form expectations about melodic continuation even without explicit formal training in music theory \cite{Rohrmeier2011,Pearce2010,Tillmann2014}.
 
For quantitative modeling, the concept of expectation requires formalization and subsequent operationalization.
In statistical-learning-based approaches, a conditional probability distribution over possible future events is estimated from musical structure and context \cite{Pearce2010}. 
Within this information-theoretic framework, the unexpectedness of an event that actually occurs can be quantified as information content, or surprisal, whereas the uncertainty of the predictive distribution before the event can be quantified as entropy \cite{Pearce2010,DiLiberto2020}. These quantities are complementary and have been associated with dissociable neural responses in their temporal dynamics \cite{DiLiberto2020}.
In this paper, musical expectation is formalized in information-theoretic terms, treating surprisal and entropy as idealized quantities that characterize unexpectedness and uncertainty. These quantities are operationalized through model-based estimates of conditional probability.
 
Consistent with the theoretical prediction/expectation framework outlined above, music ERP studies have reported multiple neural indices related to both unexpectedness- or surprisal-related processing and uncertainty- or entropy-related processing.
With respect to unexpectedness- or surprisal-related processing, several ERP components have been reported as neural markers of musical expectation violations, most prominently mismatch negativity (MMN) \cite{Friston2013,Koelsch2011,Koelsch2008,Yu2015,Brattico2006,Mencke2021,QuirogaMartinez2020}, early right anterior negativity (ERAN) \cite{Friston2013,Koelsch2011,Koelsch2008,KoelschPLosONE}, and the subsequent N5 component associated with later integration processes \cite{Friston2013,Koelsch2011,Koelsch2008,KoelschPLosONE}.
For uncertainty- or entropy-related processing, decreases in MMN amplitude with increasing Shannon entropy have been reported \cite{Lumaca2019}, and stimulus-preceding negativity (SPN) has also been linked to uncertainty processing \cite{Ono2024,Tanovic2019}.
 
At the same time, acoustic-related processing has also been characterized in ERP/EEG studies. P1/N1/P2-family effects related to the processing of pitch and time have been reported \cite{Neuhaus2008}, and naturalistic P1/N1/P2 responses to sound onsets and acoustic feature changes in real music have likewise been extracted \cite{Haumann2021}.
 
However, much of the conventional ERP evidence relies on repeated stimuli and artificial deviants, limiting the direct assessment of neural activity during continuous naturalistic music listening \cite{DiLiberto2020,Kern2023}. To address this limitation, encoding-model approaches using time-resolved regression frameworks such as the Temporal Response Function (TRF) have been applied to continuous naturalistic music \cite{DiLiberto2020,Kern2023}. Within this framework, Di Liberto et al.\ explicitly dissociated melodic expectation effects from acoustic envelope-change effects and showed that adding melodic expectation features improves the prediction of EEG and ECoG responses beyond acoustic features alone \cite{DiLiberto2020}. In this lineage of neural regression studies, expectation representations have been shown to be statistically dissociable from acoustic and adaptation effects \cite{Kern2023,DiLiberto2020,GaleanoOtalaro2024}.
 
More broadly, prior work has shown that acoustic features derived from audio or note-level information, as well as model-derived audio representations, correspond to human cortical activity \cite{Bellier2023,Tuckute2023,Mischler2025}. 
In particular, self-supervised audio models have been shown to outperform hand-engineered acoustic features and supervised speech models in predicting human cortical responses, suggesting that Artificial Neural Network (ANN)-derived audio representations capture aspects of auditory cortical processing \cite{vaidya2022selfsupervisedmodelsaudioeffectively}.
 
Building on these findings, Akama et al.\ proposed the PredANN framework based on the hypothesis that if auditory brain responses recorded under ideal conditions resemble ANN representations, then ANN-derived representations may help complement incomplete brain-response information obtained under less-than-ideal conditions. To test this idea, they trained EEG-based recognition models to predict ANN-derived teacher representations and showed improved music identification accuracy \cite{Akama2025}. However, in the original study, these teacher representations mainly reflected acoustic structure, and the role of expectation-related representations was not examined. It therefore remains unclear whether acoustic representations are optimal or whether acoustic and expectation-related representations provide complementary benefits.

\begin{figure}[!t]
    \centering
    \includegraphics[width=\textwidth]{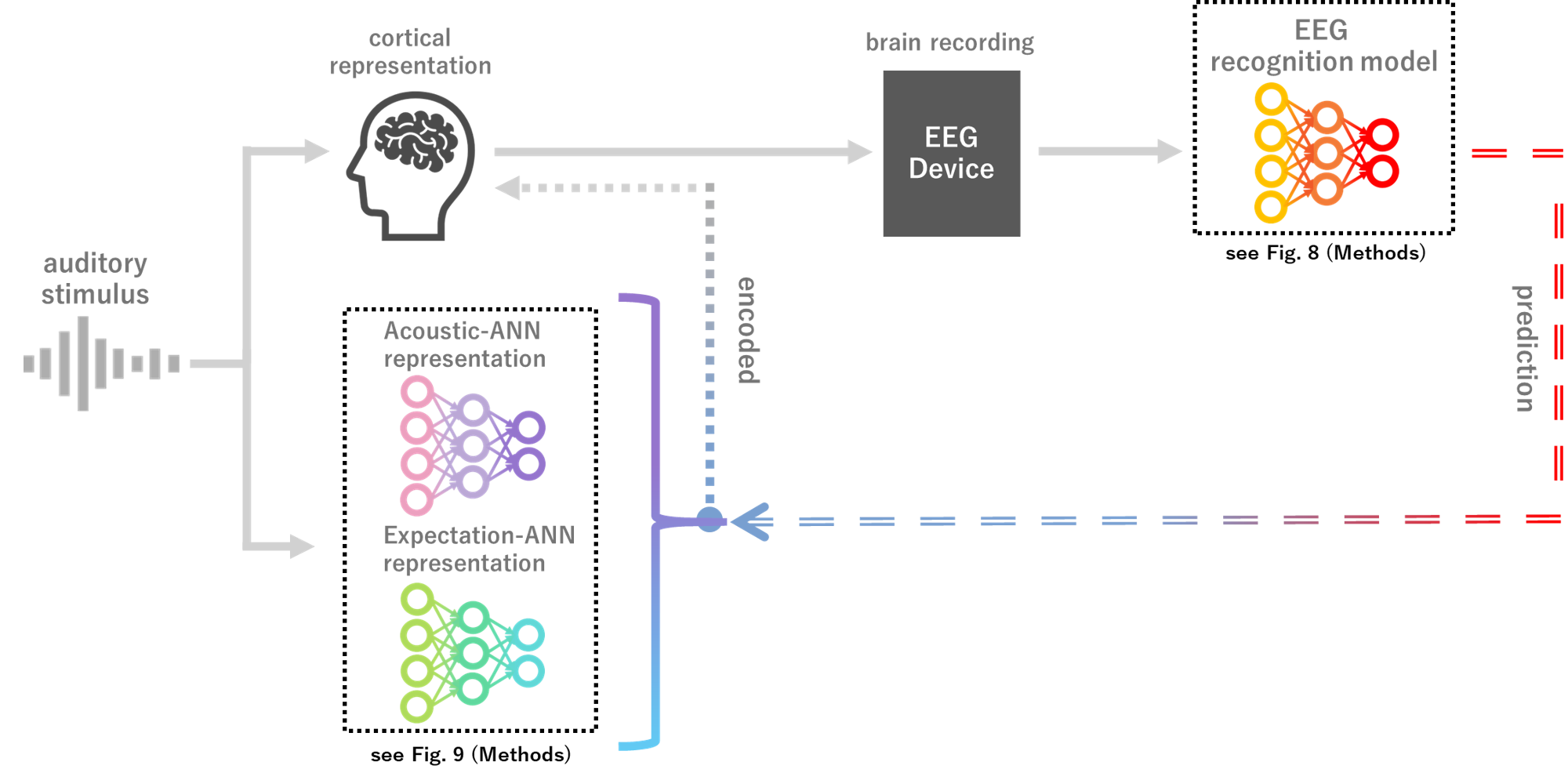}
    \caption{
\textbf{Conceptual overview of our approach.}\\
To extract representations that are encoded in the cortex, we predict corresponding ANN representations that are hypothesized to capture stimulus-related dimensions reflected in cortical activity.
As a result, the models capture different aspects of neural information. These representations can complement one another when ensembled, while each model can also individually enhance task-relevant EEG components for song ID classification.
    }
    \label{fig:concept}
\end{figure}

Here, inheriting the core hypothesis of the PredANN framework, we systematically examine how the choice of teacher representation itself influences EEG-based music identification and whether complementary effects emerge across distinct teacher types.
Specifically, we pretrain models using ANN representations computed directly from music stimuli, explicitly distinguishing acoustic representations that predominantly encode acoustic information from expectation-related representations operationalized as surprisal and entropy.
Fig.~\ref{fig:concept} illustrates the conceptual framework of our approach.
We show that each representation---acoustic, surprisal, and entropy---independently contributes to performance improvements.
Furthermore, integrating these representations yields complementary gains that surpass those achieved by simple but strong ensembles constructed solely through random initialization of the model parameters. 
The fact that recognition performance changes substantially depending on the context length used to compute expectation representations suggests that it is related to the context length of expectation processing in human music cognition. Peak recognition performance at relatively short context lengths is consistent with prior neuroscientific studies on music cognition \cite{Kern2023}. 
Together, these results establish a new framework for designing EEG recognition models grounded in the intrinsic neural encoding structure of acoustic and expectation-related information during music perception. By demonstrating that ensemble diversity can be constructed through neurobiologically distinct representations rather than initialization differences, we show that representation learning design can be guided by the organization of information encoded in cortex, thereby redefining EEG model design on a neuroscientific foundation.
This framework employs expectation features computed directly from raw signals, thereby enabling analysis of multilayer predictive structure in cortex beyond the onset- and pitch-based formulations used in prior work \cite{DiLiberto2020,Kern2023}.
Directly computed from raw signals, these expectation features do not rely on symbolic representations such as MIDI or note-level information—commonly used in prior approaches \cite{DiLiberto2020,Kern2023}—or on manual labels, allowing models to leverage large-scale unlabeled data and to extend naturally to diverse auditory stimuli.
Such label-free scalability, together with the ability to handle diverse auditory stimuli, enables training on large and heterogeneous datasets and is therefore well aligned with the direction of foundation-style EEG models capable of solving various EEG tasks.
This study provides a neuroscientifically grounded framework for EEG recognition model design that improves both performance and interpretability, with implications for neural decoding, brain–computer interfaces, and understanding predictive processing in music cognition.

\FloatBarrier
\section{Results}\label{sec:results}

Experiments were conducted using the Naturalistic Music EEG Dataset--Tempo (NMED-T) \cite{losorelli2017nmed}, comprising EEG recordings from 20 participants listening to 10 distinct full-length musical pieces. The task is formulated as a 10-way identification problem, where each EEG segment is classified into its corresponding song ID; consequently, the chance-level accuracy is 0.1.

\subsection*{Full-scratch baseline}

As a baseline, we trained a Transformer-based EEG encoder and a classification projector from scratch for the 10-class Song ID task.
The model takes 3-s EEG segments (128 channels at 125~Hz) as input and outputs a song-ID prediction.

EEG preprocessing and the time-delay setting followed the protocol described by Akama et al.\ \cite{Akama2025} (see \hyperref[sec:methods]{Methods} for details).
To assess robustness against variations in random initialization, we repeated training with three random seeds (0, 1, and 42).
The seed value 42 is a commonly used convention in computer science.
As summarized in Table~\ref{tab:baseline-seeds-en}, the full-scratch baseline achieved accuracies in the range 0.809--0.832 across seeds (mean 0.823), which we use as the baseline for subsequent comparisons.

\begin{table}[t]
  \centering
  \caption{Accuracy of the no-pretraining seed baseline across random seeds}
  \label{tab:baseline-seeds-en}
  \setlength{\tabcolsep}{24pt}
  \renewcommand{\arraystretch}{1.2}
  \begin{tabular}{lc}
    \toprule
    Seed & Accuracy \\
    \midrule
    0  & 0.832 \\
    1  & 0.809 \\
    42 & 0.827 \\
    \midrule
    Mean & 0.823 \\
    \bottomrule
  \end{tabular}
\end{table}

\subsection*{Pretraining with ANN-derived representations enhances song ID classification}
\begin{figure}[!t]
  \centering
  \includegraphics[width=0.92\linewidth]{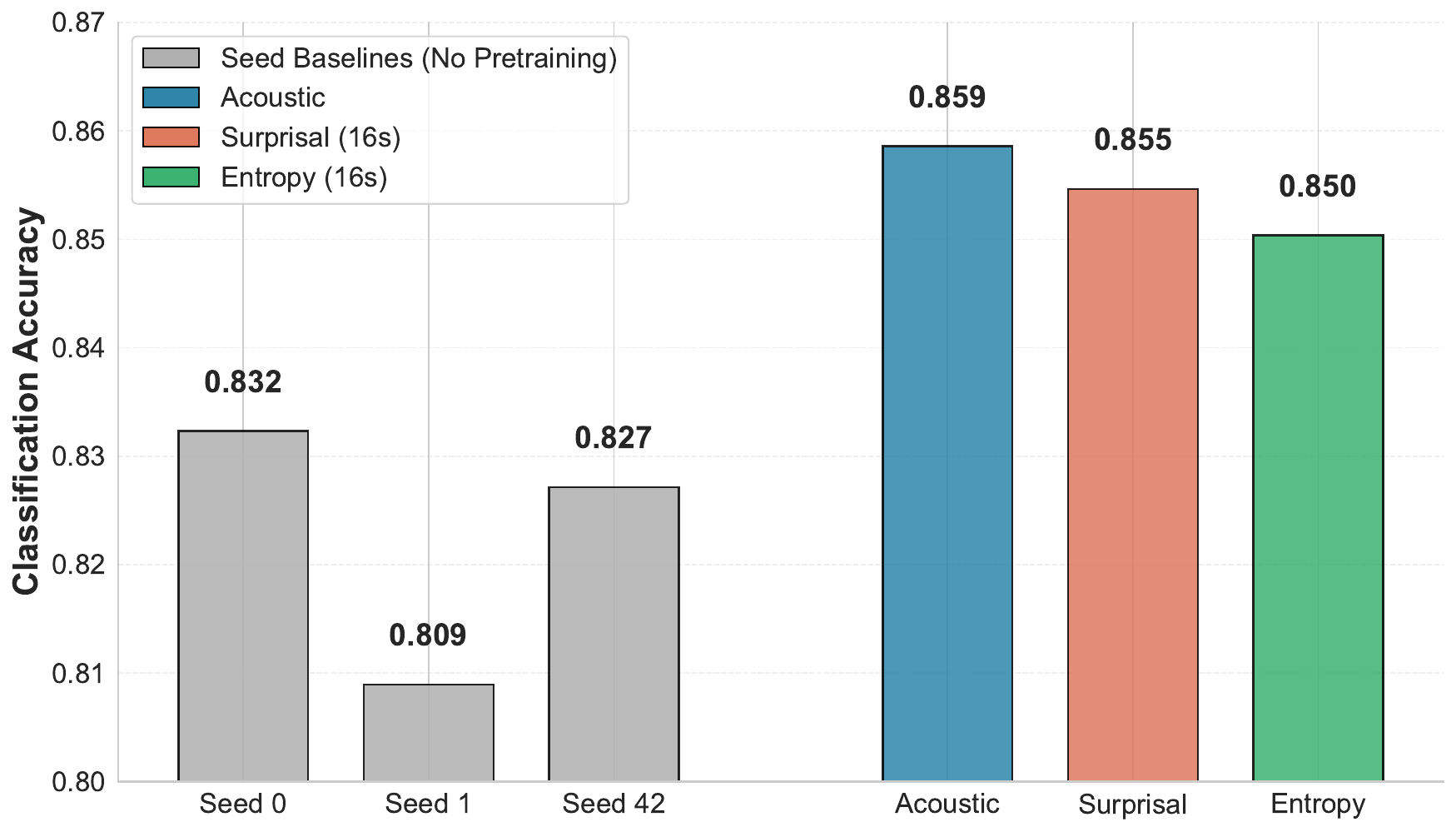}
  \caption{\textbf{Classification performance across distinct neural representations and multi-seed baselines.}\\
 Comparison of classification accuracy between individual seed baselines (trained from scratch with seeds 0, 1, and 42) and models pretrained with three distinct ANN representation types: acoustic features, surprisal (16~s context), and entropy (16~s context). All three representation-based models consistently outperformed the seed baselines, demonstrating that pretraining with ANN-derived teacher signals---whether acoustic or expectation-related---enhances EEG-based song ID classification. The acoustic model achieved the highest single-model accuracy (0.859), followed by surprisal (0.855) and entropy (0.850), each improving over the baseline mean (0.823). This pattern suggests that while all three representation types capture stimulus-related information encoded in EEG, acoustic features provide the strongest supervisory signal for this task.}
  \label{fig:single-models-en}
\end{figure}

Guided by the predictive-coding framework and evidence for dissociable cortical encoding of acoustic and expectation-related information, we investigated how the choice of target representation affects downstream Song ID classification performance.
 We compared three neurophysiologically motivated representation types as pretraining targets: Acoustic, Surprisal, and Entropy. These representations reflect complementary aspects of auditory processing---acoustic properties of the signal versus expectation-related information about unexpected events (surprisal) and uncertainty (entropy). Acoustic features were extracted using MuQ \cite{zhu2025muqselfsupervisedmusicrepresentation}, and expectation-related features (surprisal/entropy) were computed using MusicGen \cite{musicgen} (see \hyperref[sec:methods]{Methods} for details). We pretrained separate EEG encoders to predict each representation type, and then fine-tuned each of them for Song ID classification. All three representation types were evaluated under identical experimental conditions (Seed 42, 16-s context window for expectation-related features). We refer to this framework, which leverages these complementary representation types as masked prediction targets, as \textbf{\textit{PredANN++}} (full architectural details are provided in \hyperref[sec:methods]{Methods}). This setup enables the direct comparison of how representation choice impacts EEG-based decoding performance (Figure~\ref{fig:single-models-en}).

All three representation-based models outperformed the full-scratch baseline (mean accuracy 0.823 across three random seeds), suggesting that predicting ANN representations effectively completes EEG information for the song ID classification task.
Importantly, the magnitude of improvement varied across representation types: the acoustic model achieved 0.859 accuracy (a +3.6~percentage point improvement compared to the mean accuracy of full-scratch baseline), the surprisal model reached 0.855 (+3.2~pp), and the entropy model attained 0.850 (+2.7~pp). 
These results provide two key insights. First, they demonstrate that the information completion effect holds regardless of whether the target representations capture acoustic or expectation-related properties.
Second, they reveal that representation choice affects performance, with acoustic features providing the strongest supervisory signal for this task, followed by surprisal and entropy. 

\FloatBarrier

\subsubsection*{Acoustic representations: detailed analysis}

\begin{figure}[!t]
  \centering
  \includegraphics[width=0.92\linewidth]{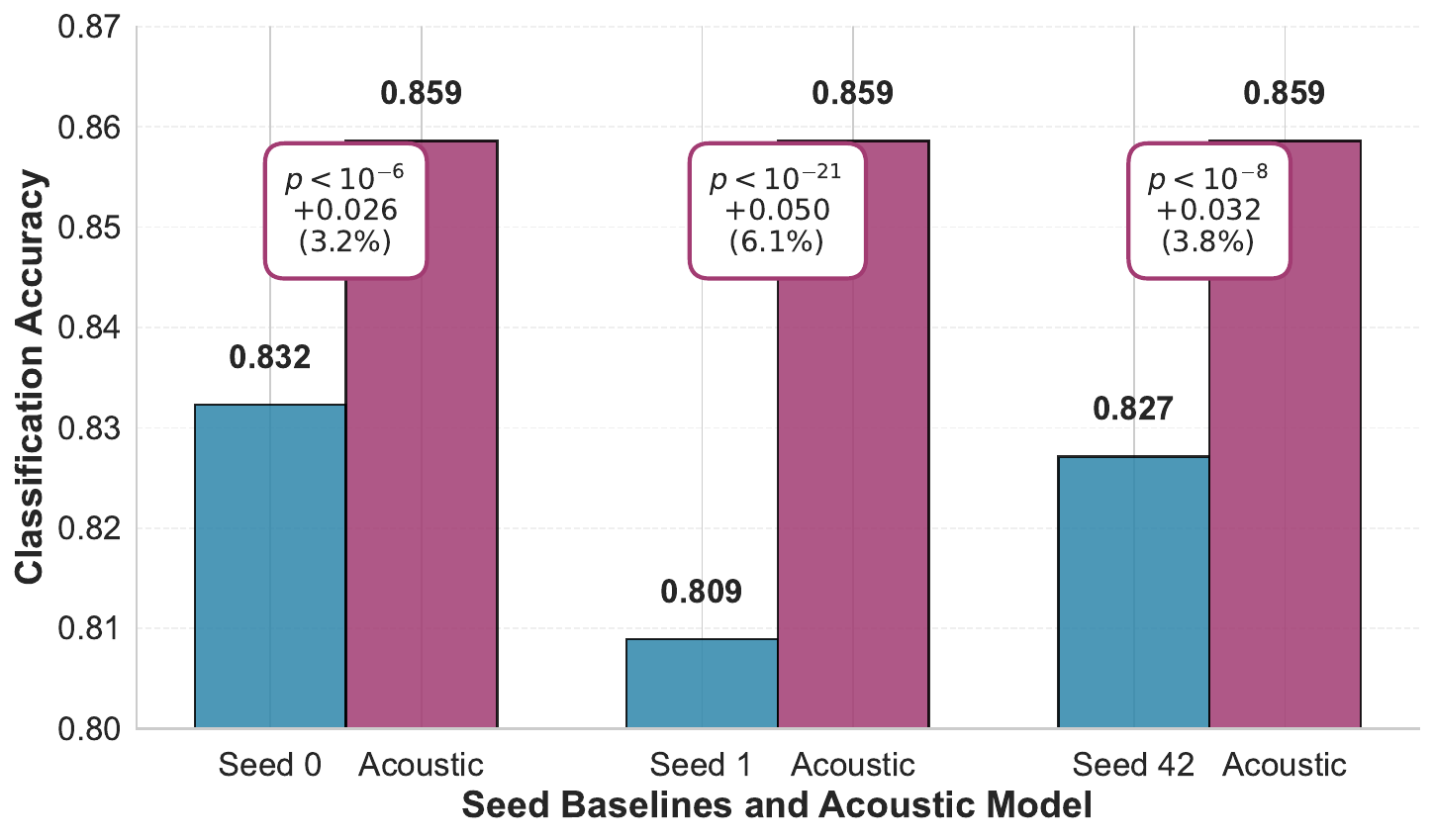}
  \caption{\textbf{Single model performance: acoustic representation pretraining vs. non-pretrained baselines.}\\ 
  Bar chart comparing classification accuracy between the acoustic feature prediction model and individual seed baselines. The acoustic model achieved 0.859 accuracy, consistently outperforming all three seed-based models: seed~0 (0.832), seed~1 (0.809), and seed~42 (0.827). Statistical annotations display McNemar's test results for each comparison, showing p-values, absolute accuracy improvements, and relative improvement percentages. All comparisons achieved $p<0.001$, confirming that acoustic pretraining provides statistically robust performance gains independent of random initialization.}
  \label{fig:muq-en}
\end{figure}

To further examine the consistency of the acoustic pretraining advantage, we conducted seed-wise statistical comparisons between the acoustic model (0.859) and each individual seed baseline using McNemar's test. As shown in Figure~\ref{fig:muq-en}, the acoustic model significantly outperformed all three seed baselines: seed~0 ($p=3.14\times10^{-7}$, $***$), seed~1 ($p=4.02\times10^{-22}$, $***$), and seed~42 ($p=1.03\times10^{-9}$, $***$), where $***$ denotes $p<0.001$. These results confirm that acoustic pretraining provides robust and statistically reliable performance gains across different random initializations.

\FloatBarrier

\subsubsection*{Context-dependent optimization of expectation-related representations}
\FloatBarrier

Using time-resolved regression (TRF), Kern et al.\ showed that neural responses to melodic surprise are best explained by models that use relatively short-range context, rather than longer context \cite{Kern2023}.

\begin{figure}[!t]
  \centering
  \includegraphics[width=0.92\linewidth]{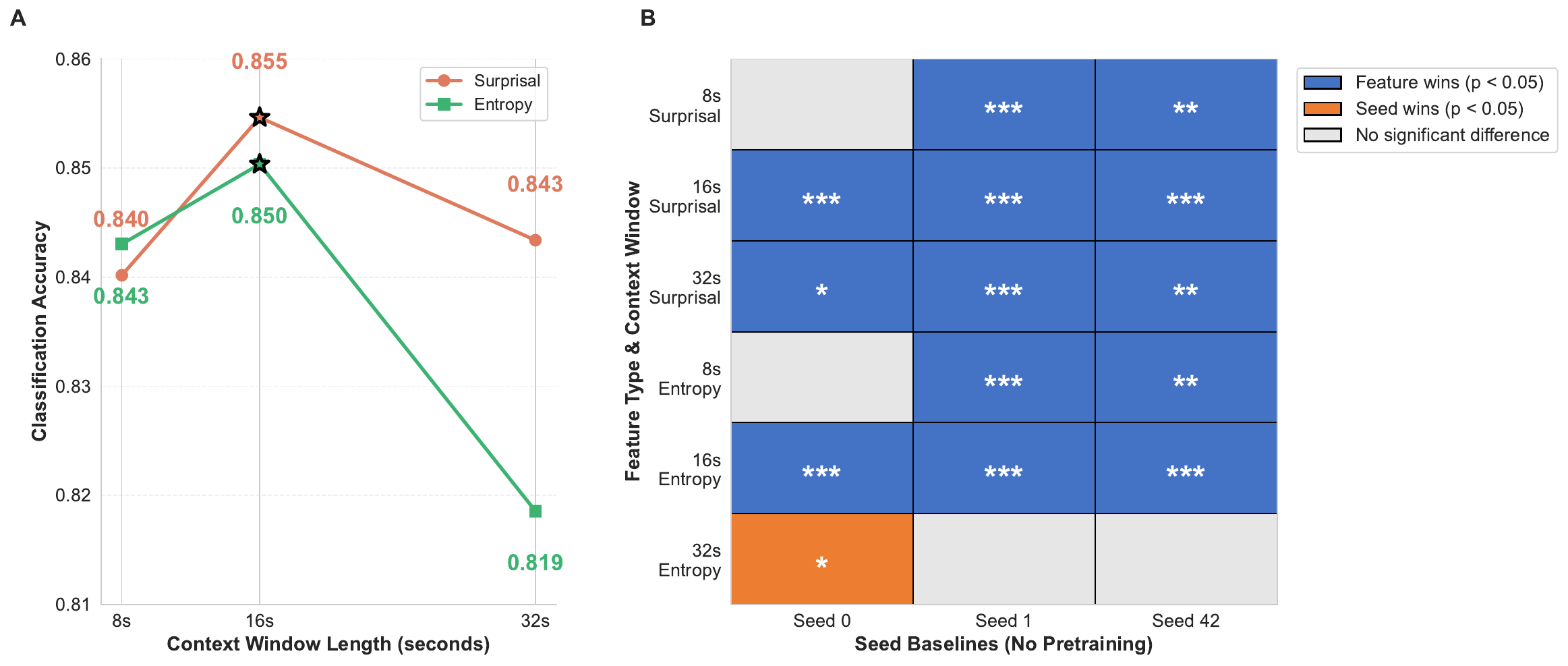}
  \caption{\textbf{Context window optimization for expectation-related features.}\\
\textbf{A} Line plot showing classification accuracy for surprisal-based (orange) and entropy-based (green) models across three context window lengths (8~s, 16~s, 32~s).
Both expectation-related features achieved peak performance at the 16-s window (surprisal: 0.855; entropy: 0.850), with star markers highlighting this optimal configuration.
Accuracy declined at both shorter (8~s) and longer (32~s) windows, suggesting that the 16-s window best captures the temporal scope of expectation processing during naturalistic music listening.
\textbf{B} Heatmap displaying McNemar's test results comparing each expectation-related model configuration (rows) against three seed baselines (columns).
Cell colors indicate statistical outcomes: blue denotes that the expectation-related model significantly outperformed the seed ($p<0.05$), gray indicates no significant difference, and orange indicates seed superiority.
Asterisks within cells denote significance levels ($*$ $p<0.05$, $**$ $p<0.01$, $***$ $p<0.001$). Only the 16-s models achieved statistical superiority over all three seeds for both surprisal and entropy.}
  \label{fig:ctx-en}
\end{figure}

To examine this context-dependency, we computed surprisal and entropy from an autoregressive music language model (MusicGen \cite{musicgen}) using three context windows: 8~s, 16~s, and 32~s. Quantitative evaluation revealed that both surprisal and entropy demonstrated peaked accuracy at 16~s context (Figure~\ref{fig:ctx-en}). Surprisal at 16~s achieved 0.855 accuracy, improving over the full-scratch baseline by +3.2~pp, with all pairwise comparisons against individual seeds reaching significance by McNemar's test ($p<0.001$). Entropy at 16~s attained 0.850 accuracy (+2.7~pp improvement), also with all seed-wise comparisons significant ($p<0.001$). In contrast, the 8-s and 32-s context windows showed inconsistent performance, with significant improvements observed only for a subset of seed baselines.
Qualitative comparisons among the audio signals, the expectation features across the three context lengths, and the EEG recordings are also provided in Figure~\ref{fig:supp3} to illustrate their temporal relationships (see also \nameref{subsec:supp3} for a detailed qualitative analysis). This overall pattern suggests that the 16-s context length optimally captures the temporal structure of cortical predictions during music listening. We therefore adopted the 16-s configuration for all subsequent ensemble experiments.

\FloatBarrier

We additionally evaluated a conservative chunk-based scheme in which MusicGen logits are computed independently within each 30-s audio segment, and surprisal and entropy are derived accordingly (see \hyperref[sec:methods]{Methods}; results are reported in Figure~\ref{fig:supp1}).

\subsection*{Deep ensembles reveal complementary benefits of diverse representations}

Prior studies have established that the human brain encodes both acoustic features \cite{Bellier2023, Tuckute2023, Mischler2025} and expectation features \cite{DiLiberto2020, Kern2023, GaleanoOtalaro2024} during music listening. Notably, the inclusion of melodic expectation features improves the prediction of neural responses beyond what is achieved by acoustic features alone \cite{DiLiberto2020}. 
Based on this potential for information complementarity, we investigated whether integrating models trained with these distinct teacher signals—Acoustic, Surprisal, and Entropy—enhances EEG-based song ID classification. Specifically, we evaluated whether ensembling these models through probability averaging yields synergistic gains over any single-feature model.

\begin{figure}[!t]
  \centering
  \includegraphics[width=0.92\linewidth]{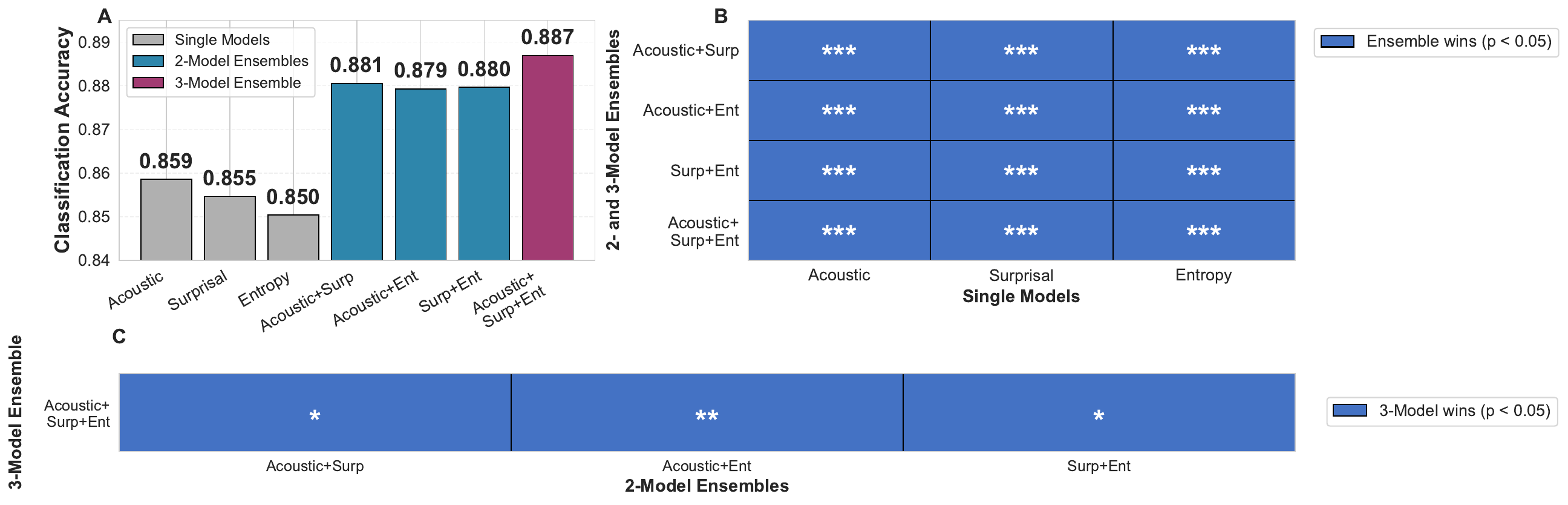}
  \caption{\textbf{Effectiveness of ensembling acoustic and expectation-related features.}\\
  \textbf{A} Bar chart comparing classification accuracy across single models, two-model ensembles, and the three-model ensemble. Accuracy increased systematically with ensemble size: single models ranged from 0.850--0.859, two-model combinations achieved 0.879--0.881, and the three-model ensemble reached 0.887, demonstrating progressive performance gains through representation integration. \textbf{B} Heatmap showing McNemar's test results comparing ensembles (rows) against constituent single models (columns). Blue cells indicate that the ensemble significantly outperformed the single model ($p<0.05$), with asterisks denoting significance levels. All ensembles achieved statistical superiority over their constituent single models, confirming synergistic benefits from combining complementary representations. \textbf{C} Heatmap displaying pairwise comparisons between the three-model ensemble (row) and all two-model ensembles (columns). The three-model ensemble significantly outperformed every two-model configuration, demonstrating that each of the three representation types---Acoustic, Surprisal, and Entropy---captures distinct information, as evidenced by their complementary contributions when integrated together.
  Statistical notation: $*$ $p<0.05$, $**$ $p<0.01$, $***$ $p<0.001$ (McNemar's test).}
  \label{fig:ensemble-en}
\end{figure}

We evaluated all possible 2-model ensembles by averaging output probabilities (Figure~\ref{fig:ensemble-en}A). Each 2-model ensemble exceeded its constituent single models by 2.0--3.0 percentage points, confirming that combining representations yields improvements. McNemar comparisons show that all 2-model ensembles significantly outperformed all three single-model baselines ($p<0.001$ for all comparisons; Figure~\ref{fig:ensemble-en}B).

The 3-model ensemble (Acoustic+Surprisal+Entropy) achieved an accuracy of 0.887.
This corresponds to a +6.4~pp improvement (7.8\% relative) over the full-scratch baseline averaged across random seeds (0.823), a +2.8~pp gain over the best single model (Acoustic, 0.859), and a +0.6~pp improvement over the best 2-model
ensemble (Acoustic+Surprisal, 0.881).

McNemar's pairwise tests confirmed that the 3-model ensemble significantly outperformed all single models (vs.\ Entropy:
$p=3.24\times10^{-23}$; vs.\ Acoustic: $p=1.69\times10^{-18}$; vs.\ Surprisal:
$p=1.39\times10^{-20}$) as well as all 2-model ensembles (vs.\ Acoustic+Entropy:
$p=0.0065$; vs.\ Entropy+Surprisal: $p=0.012$; vs.\ Acoustic+Surprisal:
$p=0.019$). Figure~\ref{fig:ensemble-en} summarizes these results, illustrating
progressive accuracy gains from single models to ensembles, and the
statistical superiority of the 3-model ensemble over single models
and 2-model ensembles.

\FloatBarrier

\subsection*{Performance of seed-based ensembling}

\begin{figure}[!t]
  \centering
  \includegraphics[width=0.92\linewidth]{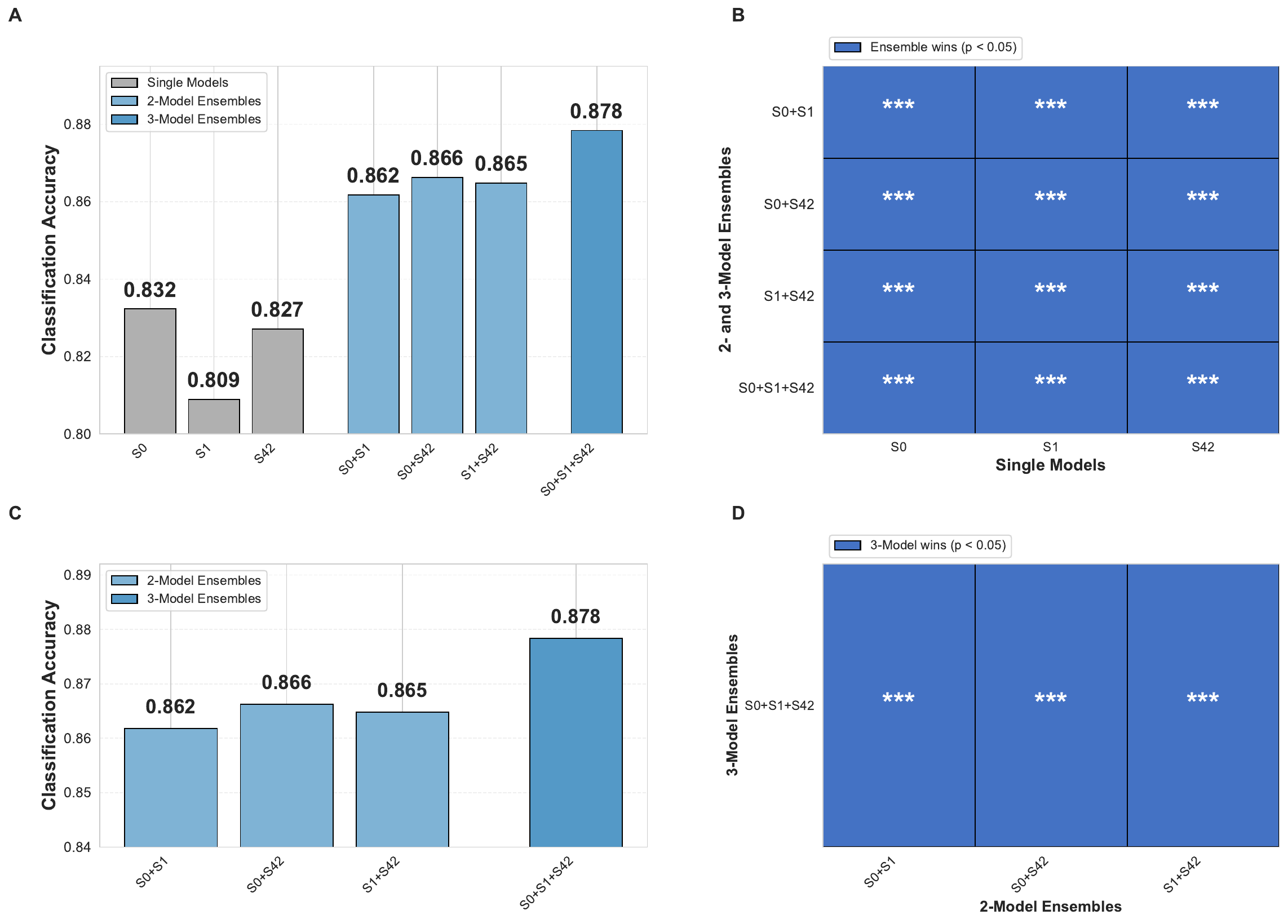}
  \caption{\textbf{Evaluation of performance gains through multi-seed deep ensembling.}\\
\textbf{A} Bar chart showing classification accuracy for individual seed models, 2-model seed ensembles, and the 3-model seed ensemble. Accuracy increased progressively from single seeds (0.809--0.832) to 2-model ensembles (0.862--0.866) to the 3-model ensemble (0.878). \textbf{B} Heatmaps displaying McNemar's test results comparing 2-model ensembles (top) and 3-model ensemble (bottom) against individual seed baselines. Blue cells indicate ensemble superiority ($p<0.05$), with asterisks denoting significance levels. All ensembles significantly outperformed individual seeds, confirming robust gains from initialization diversity. \textbf{C} Bar chart comparing 2-model and 3-model seed ensembles. \textbf{D} Heatmap showing that the 3-model ensemble significantly outperformed all 2-model configurations, demonstrating that increasing ensemble size consistently enhances performance. Statistical notation: $***$ $p<0.001$ (McNemar's test).}
  \label{fig:seed-ensemble-en}
\end{figure}

While the benefits of seed ensembling have been demonstrated in general computer vision tasks \cite{DeepEnsemble2017,DeepEnsembleTheory2019}, its effectiveness in the context of EEG-based song ID classification remains unexplored. Our results confirm that seed-based ensembling yields substantial gains in this domain (Figure~\ref{fig:seed-ensemble-en}). These robust improvements allow us to establish seed-based ensembling as a simple yet powerful baseline, against which the benefits of representation-based ensembling can be assessed in the subsequent section.

Seed-based ensembles demonstrated systematic performance improvements (Figure~\ref{fig:seed-ensemble-en}). All 2-model seed combinations achieved accuracies of 0.862--0.866, significantly outperforming individual seed models (0.809--0.832) according to McNemar's tests ($p<0.001$ for all comparisons). The 3-model seed ensemble reached an accuracy of 0.878, representing the highest performance attainable through random-initialization diversity alone. Moreover, the 3-model ensemble significantly outperformed all 2-model configurations ($p<0.001$), indicating that increasing ensemble size yields additional gains. 
Together, these results suggest that the seed-based ensemble approach provides a simple but rigorous benchmark for assessing whether representation-based diversity offers advantages beyond random initialization.

\FloatBarrier

\subsection*{Neural representation diversity outperforms initialization diversity}

Having established that both representation-based and seed-based ensembles achieve high performance in prior sections, we directly compared these two ensemble strategies. 
We specifically examined the advantage of representation diversity---integrating models trained with neurobiologically distinct teacher signals (Acoustic, Surprisal, Entropy)---over initialization diversity (seed ensembles) of equivalent size.

\begin{figure}[!t]
  \centering
  \includegraphics[width=0.92\linewidth]{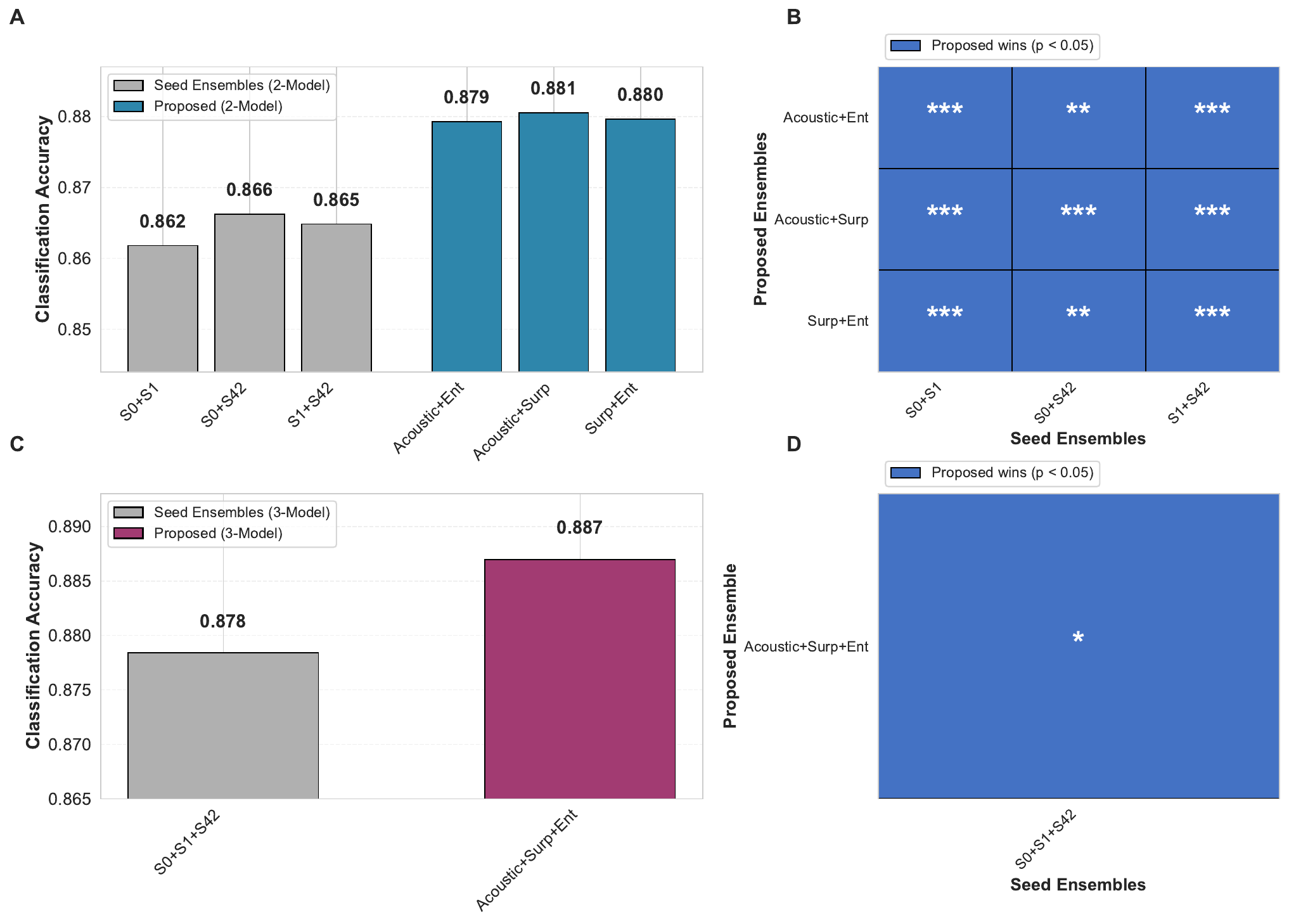}
  \caption{\textbf{Superiority of ensembling distinct representations over multi-seed ensembles.}\\
  \textbf{A} Bar chart comparing 2-model seed ensembles and 2-model representation-based ensembles. All representation-based combinations achieved higher accuracy (0.879--0.881) than seed-based ensembles (0.862--0.866). \textbf{B} Heatmap showing McNemar's test results. Blue cells indicate representation-based superiority ($p<0.05$), with asterisks denoting significance levels. All representation-based ensembles significantly outperformed all seed-based ensembles, demonstrating consistent advantages of representation diversity. \textbf{C} Bar chart comparing the 3-model seed ensemble (0.878) against the 3-model representation ensemble (0.887). \textbf{D} Heatmap confirming that the representation-based ensemble significantly outperformed the seed ensemble. The representation-based approach demonstrates that neurobiologically motivated diversity is more effective than initialization diversity alone for EEG-based song ID classification.
  Statistical notation: $*$ $p<0.05$, $**$ $p<0.01$, $***$ $p<0.001$ (McNemar's test).}
  \label{fig:seed-vs-proposed-en}
\end{figure}

Representation-based ensembles consistently achieved superior performance (Figure~\ref{fig:seed-vs-proposed-en}). In 2-model comparisons, all representation-based ensembles (Acoustic+Entropy: 0.879; Acoustic+Surprisal: 0.881; Surprisal+Entropy: 0.880) significantly outperformed seed-based 2-model versions (0.862--0.866) via McNemar's tests. The 3-model representation ensemble (0.887) also significantly outperformed the 3-model seed ensemble (0.878), exceeding the performance ceiling by +0.9~pp.

This confirms that representation diversity guided by neurobiological distinctions provides systematic advantages over conventional multi-seed averaging, establishing it as a superior ensemble strategy for EEG-based decoding.

\FloatBarrier

\section{Discussion}\label{sec:discussion}
ERP paradigms provide neurophysiological markers of expectation processing \cite{Friston2013,Koelsch2011,Koelsch2008,Yu2015}.
However, because they rely on artificially segmented stimulus sequences, these paradigms compromise stimulus naturalness and thereby constrain the study of natural music listening \cite{DiLiberto2020}.
Di~Liberto et al.\ introduced TRF modelling to partially restore stimulus naturalness.
However, their expectation features (surprisal/entropy) were derived from monophonic MIDI representations limited to pitch and onset time. 
As a result, these features remained restricted to onset-locked melodic statistics. 
Consequently, they did not define multidimensional expectation structures over continuous acoustic dimensions such as duration, dynamics, envelope shape, timbral evolution, spectral content, and reverberation, nor over higher-level attributes such as instrumentation or harmonic texture, beyond discrete MIDI symbols.

By contrast, in the present study we model predictive structure without relying on symbolic representations or explicit event segmentation. Specifically, we convert raw audio into EnCodec discrete tokens \cite{encodec} and compute surprisal and entropy from the next-token probabilities of MusicGen’s autoregressive distribution over these waveform-derived tokens \cite{musicgen} (see \hyperref[sec:methods]{Methods} for technical details). Because these next-token probabilities are defined directly on a representation trained to reconstruct the full audio signal, this formulation is holistic and multidimensional, capturing predictive regularities in natural music beyond pitch- and onset-level statistics \cite{encodec,musicgen}. To allow readers to verify how these calculated metrics correspond to actual musical transitions, we provide an interactive web-based visualization at \url{https://shogonoguchi.github.io/PredANNpp/#syncviz}. The audio examples in this demonstration are drawn from the MTG-Jamendo dataset \cite{bogdanov2019mtg}. In what follows, we interpret the improvement in EEG-based Song ID classification accuracy reported in the \hyperref[sec:results]{Results} and discuss whether our framework offers a conceptually grounded modelling strategy for representation learning, including its neuroscientific motivation, limitations, and future directions.

PredANN++ builds on Akama et al.’s hypothesis that predicting ANN representations from EEG provides an effective supervisory signal \cite{Akama2025}. Unlike the supervised ANN representations used in Akama et al., we employ self-supervised learning models, namely MuQ \cite{zhu2025muqselfsupervisedmusicrepresentation} and MusicGen \cite{musicgen}, consistent with evidence that self-supervised representations align more closely with cortical representations than supervised ones \cite{NEURIPS2022_d81ecfc8,vaidya2022selfsupervisedmodelsaudioeffectively}. We use three teacher signals: an acoustic teacher derived from MuQ embeddings, and two expectation-related teachers derived from MusicGen, namely surprisal and entropy \cite{Bellier2023,DiLiberto2020,Kern2023,Tuckute2023,GaleanoOtalaro2024,Mischler2025}. Surprisal quantifies the unexpectedness of an observed event, whereas entropy measures uncertainty in the predictive distribution prior to the event, and these quantities are distinguished as complementary information-theoretic measures \cite{Pearce2010}.

Di~Liberto et al.\ suggested that surprisal and entropy exhibit temporally complementary neural effects \cite{DiLiberto2020}. However, in a neural regression study by Kern et al., surprisal effects were more readily detected than entropy effects, and adding entropy sometimes reduced model performance \cite{Kern2023}.
One possible explanation is that this discrepancy reflects methodological constraints inherent in onset-locked regression frameworks with correlated predictors, rather than directly implying that entropy lacks neural relevance.
Uncertainty may manifest as gain-like modulation of deviance responses rather than as a strictly event-locked main effect \cite{Lumaca2019}.
Because surprisal is defined after an event and is therefore event-specific, whereas entropy reflects pre-event uncertainty in the predictive distribution, entropy may be difficult to capture using predictors computed only at note onsets. Moreover, surprisal and entropy are themselves correlated \cite{DiLiberto2020}, and incorporating correlated quantities within a single regression model can hinder reliable estimation of their unique contributions. Consistent with this concern, Galeano-Otálvaro et al.\ demonstrated entropy’s neural contribution by removing individual predictors from a full TRF model that included both acoustic and melodic predictors \cite{GaleanoOtalaro2024}. For these reasons, we avoid forcing surprisal and entropy to compete within a single shared representation and instead pretrain separate encoders using each quantity as an independent teacher signal.

\begin{figure}[ht]
    \centering
    \includegraphics[width=0.95\linewidth]{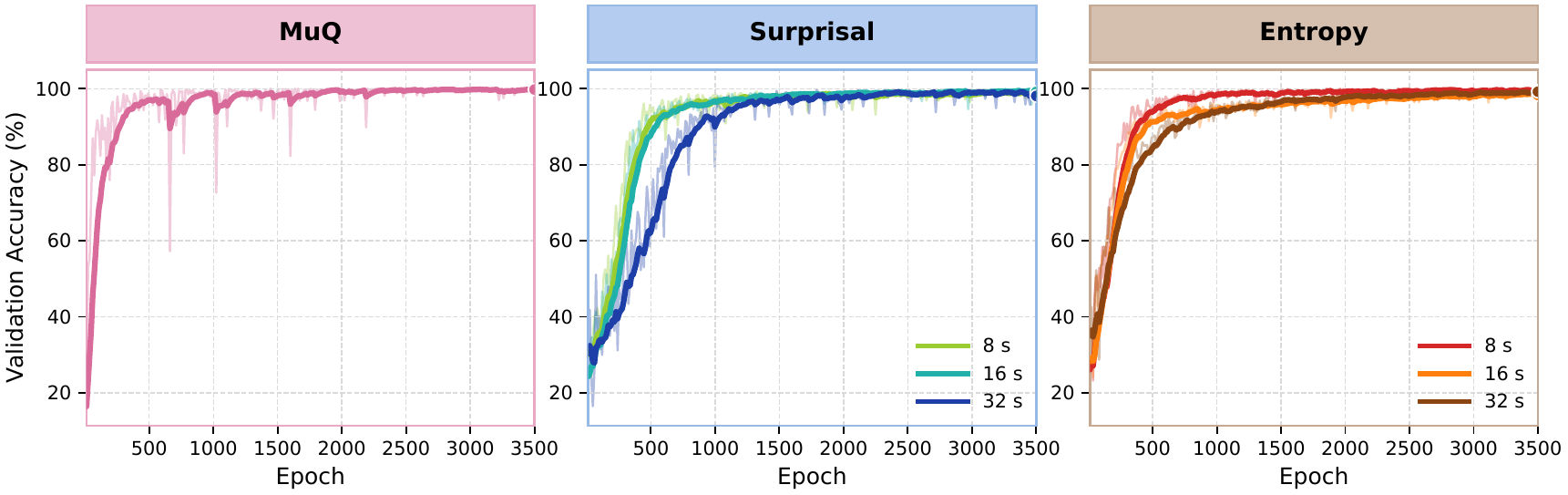}
    \caption{
        \textbf{Validation accuracy across training epochs for MuQ, Surprisal, and Entropy representations when used directly for Song ID classification.}\\
        Each panel shows validation accuracy (\%) as a function of training epoch for models that directly use discretized music-feature sequences as input, without EEG signals. For each 3-second segment, the input consists solely of discretized feature tokens: MuQ (75 steps at 25~Hz), Surprisal (150 steps at 50~Hz), or Entropy (150 steps at 50~Hz). Each scalar discrete value (0--127) is projected to a 512-dimensional embedding through a linear layer (1$\rightarrow$512). A learnable \texttt{[CLS]} token is prepended, and learnable positional embeddings are added. The sequence is processed by a 2-layer Transformer encoder (embedding dimension 512, 8 attention heads, MLP ratio 4.0, GELU activation, dropout 0.1), identical to the Transformer architecture used in the main EEG recognition experiments. The hidden representation of the \texttt{[CLS]} token is passed through a classification head composed of a linear layer (512$\rightarrow$256), batch normalization, ReLU activation, and a final linear layer (256$\rightarrow$10) to predict Song ID. The batch size is fixed to $B=48$. Optimization is performed using Adam. The learning rate is 0.003 for MuQ-based modeling and 0.0003 for Surprisal- and Entropy-based modeling, consistent with the learning-rate settings used in the main experiments. All curves represent exponential moving averages of validation accuracy. Under this modeling setup, all three representations independently achieve validation accuracies approaching 100\%, demonstrating that each representation alone contains sufficient information to almost perfectly solve the 10-class Song ID task.
    }
    \label{fig:supp2}    
\end{figure}

Prior neural regression studies suggest a relative spatiotemporal segregation between lower-level sensory responses and melodic-expectation responses, rather than a complete anatomical double dissociation.
Temporally, Di~Liberto et al.\ reported melodic-expectation effects centered around $\sim 200$\,ms compared with an acoustic-envelope component around $\sim 50$\,ms \cite{DiLiberto2020}, whereas Kern et al.\ reported earlier note-onset responses around $\sim 75$\,ms and later melodic-surprise responses around $\sim 200$\,ms and 300--500\,ms \cite{Kern2023}.
Spatially, Di~Liberto et al.\ found larger expectation effects in planum temporale than in Heschl's gyrus \cite{DiLiberto2020}, whereas Kern et al.\ reported source-model evidence that onset-related activity was best explained by bilateral Heschl's gyri, while surprise-related activity was best explained by slightly more lateral sources encompassing both bilateral Heschl's gyri and bilateral superior temporal gyri \cite{Kern2023}.
Taken together, these findings suggest that, compared to acoustic processing, expectation-related processing involves relatively more distributed temporal-cortical regions beyond Heschl's gyrus.
This relative spatiotemporal segregation suggests that denoising aligned with acoustic and expectation signals can complement information across both time and cortical space.
In practice, we train an EEG encoder $F$ and decoder $G$ such that $G(F(x))$ approximates a stimulus-derived music feature $m$, thereby encouraging $F$ to learn representations predictive of $m$.
A natural question is whether aligning EEG representations with $m$ benefits Song ID classification. 
To test this directly, we trained a Transformer classifier to predict Song ID from acoustic, surprisal, or entropy sequences alone, without EEG.
Figure~\ref{fig:supp2}  shows that acoustic features, surprisal, and entropy each achieve near-perfect accuracy (approaching $100\%$) when used directly.
Therefore, if these features are successfully recovered from EEG, they are sufficient for Song ID classification.

We further adopt a deep-ensemble strategy by averaging class probabilities from encoders pretrained with different teacher signals. Deep ensembles improve performance when individual models produce less correlated errors \cite{DeepEnsemble2017,DeepEnsembleTheory2019}. Unlike seed ensembles, which vary only random initialisation while keeping architecture, objective, and teacher fixed, our ensemble combines models pretrained with distinct teacher representations---acoustic features, surprisal, and entropy. These distinct teachers induce different inductive biases in the encoder $F$, encouraging more diverse error patterns and yielding gains beyond conventional seed-based ensembles.

We next discuss the neuroscientific relevance of the context window used by MusicGen to compute surprisal and entropy. Kern et al.\ showed that EEG/MEG regression from a Music Transformer was most accurate when conditioning on fewer than roughly ten notes ($\sim$2--4\,s) \cite{Huang2019MusicTransformer}. In our study, varying the context window $W \in \{8,16,32\}$ revealed a clear peak at $W=16$. Although the tasks differ and direct comparison is not possible, the fact that performance is maximised at relatively short windows in both cases indicates a shared non-monotonic dependence on context length, supporting the neuroscientific plausibility of our formulation.

Several factors may explain why a longer window is optimal here.
First, unlike the monophonic MIDI-synthesised stimuli used by Kern et al., our stimuli are commercially produced polyphonic songs in which predictive structure may extend beyond ten-note contexts through rhythm, dynamics, timbre, texture, and other longer-timescale musical regularities.
Second, unlike Di~Liberto et al.\ and Kern et al., who computed surprisal only at discrete note-onset positions derived from MIDI, our surprisal is computed at every 20-ms frame from a waveform-token language model. This frame-wise computation allows predictive quantities to be defined over the entire continuous time series rather than only at note events. As a result, our formulation can capture expectation-related structure that is not strictly tied to discrete note onsets, including slow dynamics and sub-note acoustic variations. A representation-learning objective that does not enforce strict onset locking may therefore better exploit these continuous predictive signals \cite{Kern2023}. Third, because our surprisal and entropy are computed from a waveform-token language model (MusicGen over EnCodec tokens \cite{musicgen,encodec}), they may reflect expectations not only for pitch/onset-like events but also for duration, dynamics, timbral evolution, and instrumentation changes unfolding over longer timescales. 
Importantly, although our encoder processes only a 3-s EEG segment, the teacher signals themselves are computed from longer context windows. 
By pretraining the model to predict these long-context quantities, the encoder is encouraged to extract local neural patterns that reflect predictive structure shaped by extended stimulus history.
In our setting, the Transformer-based encoder has strong representation capacity and the ability to aggregate information over longer ranges, which may facilitate learning effective representations from such complex predictive structure.

We next discuss limitations and future directions, focusing on surprisal/entropy computation and dataset requirements. 
In this study, we used MusicGen \cite{musicgen} as the music language model to compute surprisal and entropy directly from audio. Our first criterion in selecting the model was that it should reflect long-term statistical regularities acquired from a large-scale music corpus, because expectation features derived from such long-term learning better explain cortical responses than those based only on short-term regularities \cite{Kern2023}. In this line of work, AudioLM established the general framework of discrete audio language modeling without symbolic representations, while Jukebox, MusicLM, and MusicGen provided music-specific raw-audio alternatives trained at large scale \cite{AudioLM,Jukebox2020,MusicLM2023,musicgen}.
We selected MusicGen because it offers a single-stage autoregressive Transformer over EnCodec tokens, rather than a hierarchical multi-stage generation pipeline \cite{musicgen,encodec}. 
This makes next-token probabilities straightforward to extract from a single model operating on a single tokenization scheme, while still leveraging a large licensed music corpus and strong music-generation quality \cite{musicgen}.
For these reasons, MusicGen provided the most practical and conceptually clean basis for computing frame-wise surprisal and entropy in the present study.

MusicGen is trained as an autoregressive language model over four discrete EnCodec Residual Vector Quantization (RVQ) codebooks per time step \cite{musicgen,encodec}.
Let \(q_t^{(k)}\) denote the token at physical time \(t\) in codebook \(k \in \{1,2,3,4\}\). 
In principle, the surprisal of the full EnCodec representation at physical time \(t\) is the joint quantity
\[
-\log p_{\theta}\!\left(q_t^{(1)}, q_t^{(2)}, q_t^{(3)}, q_t^{(4)} \,\middle|\, Q_{<t}\right),
\]
where \(Q_{<t}\) denotes all RVQ tokens at earlier physical times.
We do not compute this full joint quantity directly because the publicly released MusicGen model is trained with the delay interleaving factorization across RVQ codebooks \cite{musicgen}. 
To address this constraint, we define surprisal and entropy for the target token \(q_t^{(1)}\), namely the codebook-1 token at physical time \(t\), while conditioning on the delay-pattern context made available by the released MusicGen implementation.
This choice does not restrict the conditioning context to codebook 1; it specifies only that the predictive quantity is evaluated for the \(k1\) target token.
Among the four codebooks, \(q_t^{(1)}\) is the only target token whose predictive quantity can be defined without conditioning on same-step or future-step information relative to that token.
Under the delay pattern, the conditioning context for \(k3\) and \(k4\) can contain tokens from later physical time steps in other codebooks, which introduces future reference. 
Although \(k2\) avoids this stronger form of leakage, \(q_t^{(2)}\) is still conditioned on \(q_t^{(1)}\) from the same physical time step and is therefore not strictly pre-event. 
For this reason, we compute surprisal and entropy from the implemented delay-pattern conditional probability \(p_{\theta}\!\left(q_t^{(1)} \mid C_t^{\mathrm{delay}}\right)\).
A separate issue is that the delay interleaving schedule does not always include all tokens from earlier time steps in the conditioning context used to predict \(q_t^{(1)}\).
In particular, some tokens from the immediately preceding step can be absent from the context due to the delayed interleaving order of codebooks. 
For example, the token \(q_{t-1}^{(2)}\) may be missing from the conditioning context even though it belongs to the past (\(t-1 < t\)). 
We therefore treat the implemented probability as an operational approximation to the conditional probability that would be obtained under an exact flattening factorization with the full past context,
\[
p_{\theta}\!\left(q_t^{(1)} \mid C_t^{\mathrm{delay}}\right) \approx p_{\theta}\!\left(q_t^{(1)} \mid C_t^{\mathrm{full}}\right),
\]
where \(C_t^{\mathrm{full}}\) denotes all past RVQ tokens across codebooks.
This approximation is practically reasonable because delay-pattern MusicGen preserves high-quality generation while achieving a computationally efficient inexact factorization \cite{musicgen}. 
In RVQ, each quantizer encodes the residual error left by the previous stage, and prior RVQ-based work characterizes the first codebook as the most important component of the representation \cite{musicgen,encodec}. 
Focusing on \(k1\) therefore yields a causally valid predictive quantity while targeting the dominant signal structure.

As a complementary direction, future work could employ an acoustic language model trained with an exact flattening factorization across all RVQ codebooks. This would allow computation of joint probabilities over the full multi-codebook representation without relying on the delay-pattern approximation.
Furthermore, if future studies establish that different codebooks capture disentangled musical attributes such as melody, rhythm, harmony, dynamics, or timbre \cite{Wang2022DRL}, then codebook-specific surprisal and entropy could be interpreted as attribute-specific predictive quantities, enabling finer-grained analysis of musical expectation.

We next describe dataset-related limitations and future directions. Although the framework adopts a pretrain--fine-tune paradigm, both stages depend on the same dataset. To support broad generalisation from pretraining, the EEG corpus used for pretraining must be substantially expanded. Moreover, NMED-T includes limited numbers of songs and participants, so subject-independent generalisation and transfer to external EEG datasets remain untested. 

Addressing these issues will require high-quality, large-scale EEG datasets and some standardisation of electrode configurations. Importantly, our surprisal and entropy representations do not rely on symbolic labels such as MIDI or on manual annotation, and can therefore be computed directly from general music audio. 
More broadly, the same discrete token-based formulation is, in principle, applicable to audio beyond music, including speech and environmental sounds. With such resources, the approach could move beyond task- or domain-specific EEG decoders and connect to EEG foundation models that generalise across domains via self-supervised auditory representations.

Overall, PredANN++ provides a framework for designing EEG recognition models grounded in the intrinsic neural representational structure of acoustic and expectation-related information encoded in cortex during music listening.
The improvements in Song ID decoding achieved by Acoustic, Surprisal, and Entropy teachers, together with the fact that their ensemble surpasses strong seed-based ensembles, suggest complementary structures of acoustic and expectation information in cortex, as well as complementary roles of surprisal and entropy. 
These findings further demonstrate that effective representation-learning design can be guided by the organization of information encoded in the brain.
Because expectation features are computed directly from raw audio, the proposed framework does not rely on symbolic representations or manual annotations and enables principled investigation of multilayer musical expectation structures under natural listening conditions.
This design naturally extends to diverse auditory stimuli and aligns with the direction of foundation-style EEG models capable of addressing multiple EEG tasks through heterogeneous data.
Together, this study provides a neuroscientifically grounded principle for EEG recognition model design, with implications for neural decoding and its applications in brain–computer interfaces, and for understanding predictive processing in music cognition.
\section*{Methods}
\phantomsection
\label{sec:methods}
\begin{figure}[t]
    \centering
    \includegraphics[width=\textwidth]{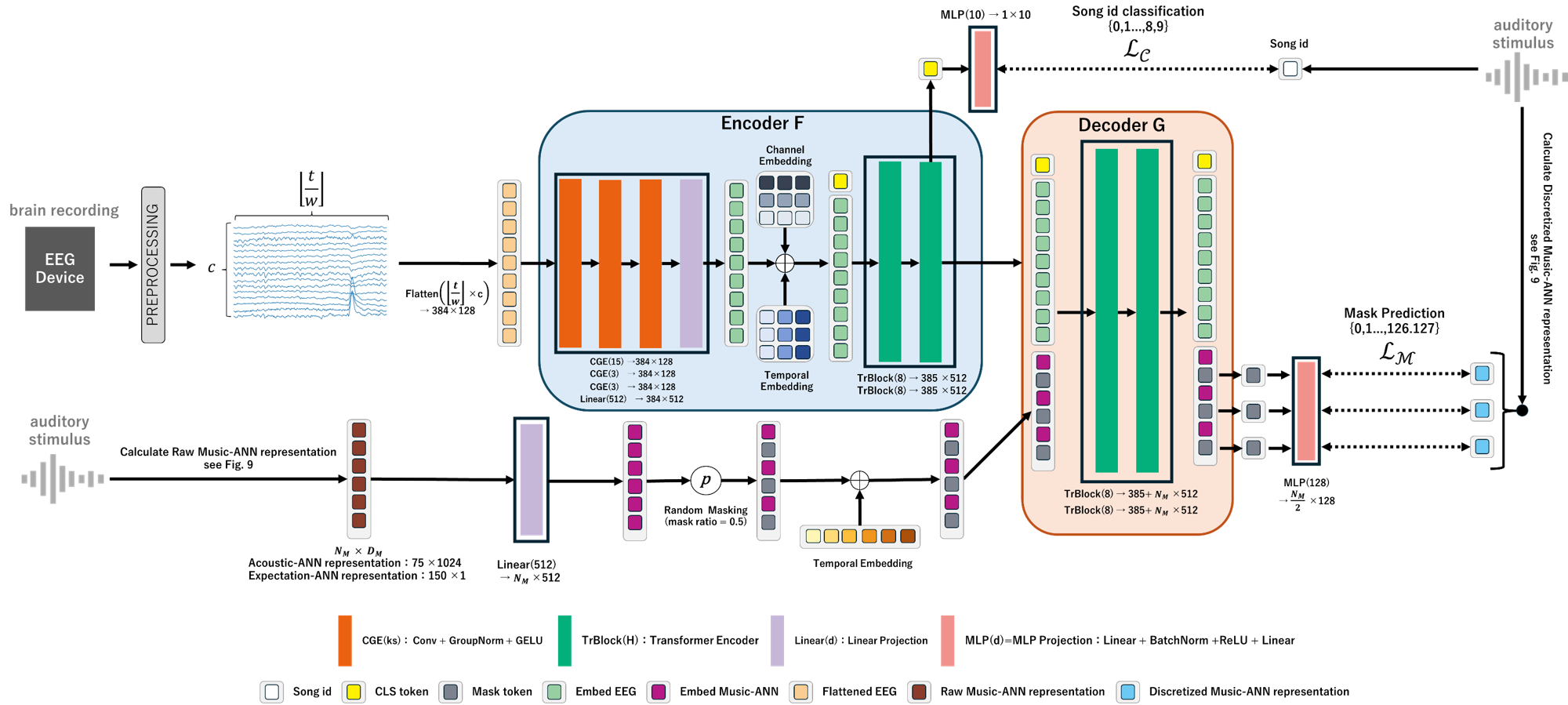}
\caption{
\textbf{Architecture of PredANN++.}\\
An EEG encoder $F$ maps a 3-s EEG segment to a latent representation using temporal patch embedding and a Transformer.
During pretraining, a decoder $G$ predicts masked tokens of discretized music-derived teacher representations (Acoustic, Surprisal, or Entropy), while $F$ is jointly regularized by an auxiliary Song ID classification loss.
After pretraining, $G$ is discarded and $F$ is fine-tuned for EEG-based Song ID classification.
}
    \label{fig:architecture}
\end{figure}

\subsection*{Study overview}

We propose \textit{PredANN++}, a pretraining framework for EEG-based song ID classification that learns EEG representations predictive of discretized music features derived from auditory stimuli. As illustrated in Fig.~\ref{fig:architecture}, the model adopts an encoder--decoder architecture consisting of an EEG encoder $F(\cdot)$ and a decoder $G(\cdot)$.

The primary objective of the pretraining stage is to guide the encoder to extract EEG representations that preserve stimulus-related information encoded in music-derived teacher signals. Specifically, given a 3-s EEG segment $x$, the encoder--decoder pathway is trained such that the decoded representation $G(F(x))$ can predict masked tokens of a discretized music feature sequence $m$ computed from the corresponding audio stimulus. 
Inspired by the Supervised Masked Autoencoders (SupMAE) framework \cite{supmae}, Song ID classification from $F(x)$ is incorporated concurrently as an auxiliary objective. While the nominal loss weight for the classification task is set higher than that of the masked prediction task, it is important to note that the prediction loss is computed over a dense sequence of numerous masked tokens per segment. Consequently, frame-level predictive signal dominates the overall learning dynamics. The sequence-level classification objective serves to guide and stabilize the representation learning process rather than acting as the primary driver of pretraining.

The teacher signal $m$ always consists of discretized music features. 
These features are obtained from two distinct audio-based models: (i) MuQ features derived from a masked language modeling-based self-supervised audio model and discretized using $k$-means clustering, and (ii) surprisal and entropy values computed from an autoregressive audio language model and discretized using quantile binning (see Fig.~\ref{fig:ann_representation}, \nameref{subsec:supp1} and \nameref{subsec:supp2} for details of the processing pipeline).

After pretraining, the decoder $G(\cdot)$ and the music-feature prediction heads are discarded. The encoder $F(\cdot)$ is then fine-tuned using only the Song ID classification objective. This two-stage training strategy allows the encoder to first acquire EEG representations aligned with specific stimulus-related dimensions and subsequently adapt them to the downstream song ID classification task.

The intuition underlying this design is as follows. Let $x$ denote an EEG segment and $m$ a discretized music feature sequence derived from the corresponding auditory stimulus. By enforcing that $G(F(x))$ accurately predicts masked elements of $m$, the encoder $F(\cdot)$ is encouraged to extract EEG representations that preserve information relevant to the stimulus attributes captured by $m$.

In this study, we explicitly distinguish between two classes of music representations. Acoustic representations consist of MuQ features and primarily encode acoustic information present in the audio signal. Expectation features consist of surprisal- and entropy-based representations and reflect properties of auditory processing related to musical expectations. Entropy explicitly quantifies uncertainty in the predictive distribution, while both surprisal and entropy capture expectation-related information derived from autoregressive next-token probabilities.

Our framework is designed to integrate these diverse representations, leveraging the finding that the human brain encodes both acoustic \cite{Bellier2023, Tuckute2023, Mischler2025} and expectation \cite{DiLiberto2020, Kern2023, GaleanoOtalaro2024} information. Since the inclusion of expectation features has been shown to improve the prediction of neural responses beyond what is achieved by acoustic features alone \cite{DiLiberto2020}, we employed an ensemble strategy to combine encoders pretrained with these different teacher signals. By integrating these complementary representations, our approach aims to capture the diverse information axes, thereby enhancing the performance and robustness of EEG-based song ID classification (see \hyperref[sec:discussion]{Discussion} for a detailed analysis of ensemble effects).

\subsection*{Dataset and preprocessing}

We used the Naturalistic Music EEG Dataset--Tempo (NMED-T) \cite{losorelli2017nmed}, a publicly available dataset consisting of EEG recordings acquired while participants listened to naturalistic music. The dataset includes EEG data from 20 participants who listened to 10 commercially released musical pieces. Neural activity was recorded from 128 scalp electrodes during continuous music listening.

EEG signals in the original dataset were recorded at a sampling rate of 1000~Hz. Following the preprocessing pipeline provided by Losorelli et al.\ \cite{losorelli2017nmed}, the raw EEG signals were high-pass filtered at 0.3~Hz, notch filtered at 59--61~Hz to remove line noise, and low-pass filtered at 50~Hz. After filtering, the signals were downsampled to 125~Hz. We used the preprocessed EEG signals provided by the dataset without modifying the original filter settings, ensuring consistency with prior analyses of the NMED-T dataset.

Unless otherwise noted, the dataset configuration, data splits, and EEG preprocessing procedures used in this study closely follow those described in Akama et al.\ \cite{Akama2025}. This design choice facilitates direct and controlled comparisons of Song ID classification performance between the proposed framework and the original PredANN method described therein.

\subsection*{EEG preprocessing and segmentation}

All EEG signals were analyzed at a sampling rate of 125~Hz with 128 scalp channels, following the preprocessing provided with the NMED-T dataset \cite{losorelli2017nmed}. To ensure consistency across recordings, all EEG recordings were truncated to a common maximum duration of 4~min. Each 4-min recording was then segmented into non-overlapping 30-s excerpts, resulting in eight excerpts per musical piece.

The 30-s excerpts were divided into training and validation sets using a 75:25 split. To preserve the original song distribution in both sets, stratified sampling was applied with respect to Song ID labels. Specifically, we used the \texttt{train\_test\_split} function from the \texttt{sklearn.model\_selection} module, specifying the \texttt{stratify} parameter as the song label \cite{scikit}. To guarantee reproducibility, the random seed was fixed to 42, consistent with the experimental setup of Akama et al.\ \cite{Akama2025}.

Based on prior neurophysiological findings regarding temporal delays in auditory cortical responses reflecting melodic expectations \cite{DiLiberto2020}, a fixed temporal shift of 200~ms was applied to align EEG signals with the corresponding auditory stimuli. Concretely, the starting point of each EEG segment was shifted forward by 200~ms, which corresponds to 25 samples at a sampling rate of 125~Hz. This temporal alignment is consistent with evidence that higher-order auditory cortical responses to music typically emerge with latencies in the range of approximately 150--250~ms \cite{DiLiberto2020}. Importantly, this delay value was not tuned in the present study, but was adopted directly from Akama et al.\ \cite{Akama2025}, who systematically compared multiple latency conditions and demonstrated that a delay of approximately 200~ms yielded the highest song ID classification accuracy.

The input to the neural network consisted of 3-s EEG segments, corresponding to 375 samples at 125~Hz. For both training and evaluation, we first constructed sliding windows within each 30-s excerpt using a window size of 1000 samples (8~s) and a stride of 200 samples (1.6~s). From each 8-s window, a single 3-s EEG segment was extracted. During training, the starting position of the 3-s segment was sampled uniformly at random from the valid range within each window, whereas during evaluation, the segment was extracted from the temporal center of the window to ensure deterministic and reproducible evaluation.

EEG normalization was performed independently for each 3-s EEG segment and for each channel. Specifically, a \texttt{RobustScaler} \cite{scikit} based on the median and interquartile range was applied to the time series of each channel, followed by clamping the scaled values to the range $[-20,\,20]$. This normalization strategy suppresses the influence of transient large-amplitude noise and artifacts while preserving the relative temporal structure of the EEG signals. Because normalization was applied independently to each sample, no statistics were shared across training, validation, or test sets, thereby preventing information leakage. Implementation details are provided in the function \texttt{normalize\_EEG\_4} in the file \href{https://github.com/ShogoNoguchi/PredANNpp/blob/main/codes_3s/predann/datasets/preprocessing_eegmusic_dataset_3s.py}{\texttt{preprocessing\_eegmusic\_dataset\_3s.py}}.

\begin{figure}[t]
    \centering
    \includegraphics[width=\textwidth]{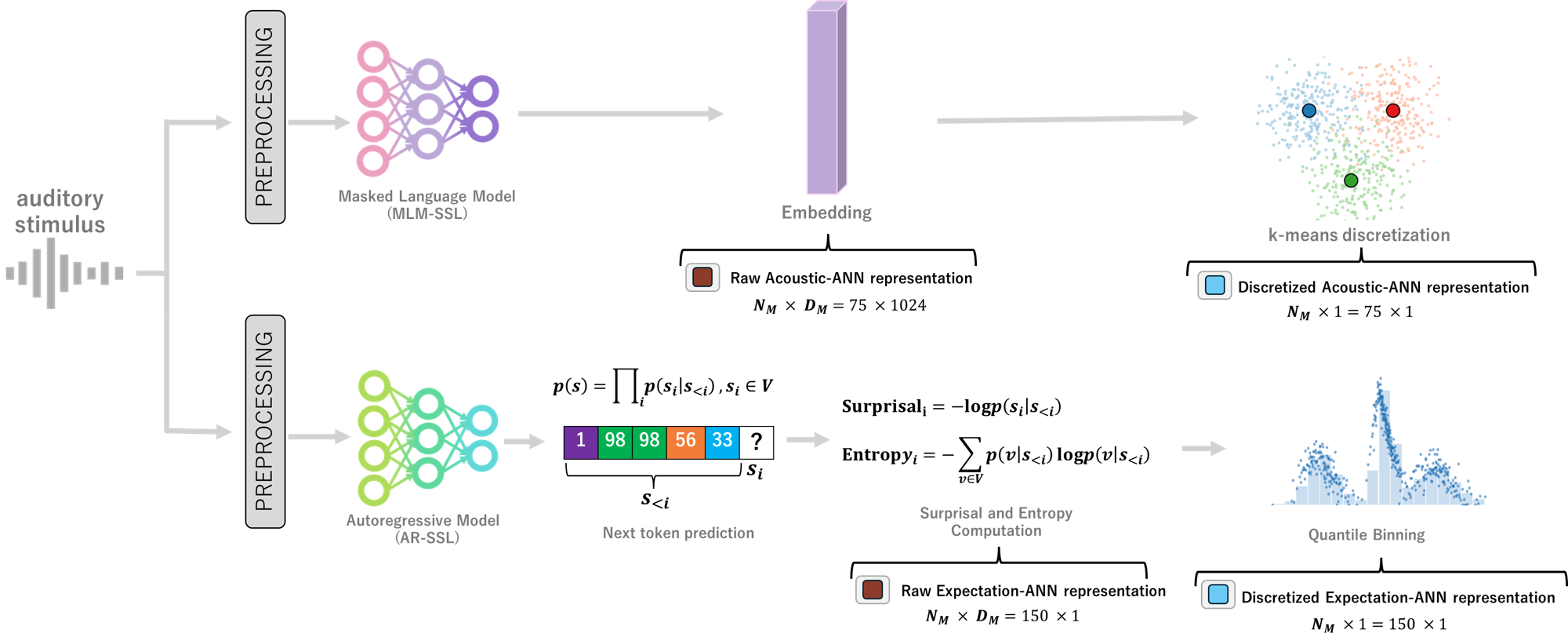}
\caption{
\textbf{ANN-representation calculation and discretization from audio.}\\
Acoustic representations are extracted using a masked language model (MLM-SSL; MuQ),
while expectation-related representations are computed using an autoregressive model (AR-SSL; MusicGen)
as surprisal and entropy.
The resulting continuous (raw) representations are used as inputs to the decoder $G$ during masked pretraining,
whereas their discretized counterparts (MuQ: $k$-means; Surprisal/Entropy: quantile binning)
serve as discrete teacher targets for prediction.
All representations are temporally aligned to 3-s EEG segments.
}
    \label{fig:ann_representation}
\end{figure}

\subsection*{ANN representations from audio}

We constructed three types of artificial neural network (ANN) representations directly from the acoustic signals, as summarized in Figure~\ref{fig:ann_representation}. These representations were designed to capture complementary aspects of auditory processing and were used as teacher signals for masked-prediction-based multitask pretraining. 
Specifically, we extracted (i) an acoustic representation derived from a self-supervised music foundation model \cite{zhu2025muqselfsupervisedmusicrepresentation}, and (ii) expectation-related representations derived from surprisal and entropy computed from an autoregressive music language model \cite{musicgen}.
All three ANN representations were discretized into 128 levels (0--127) and served as prediction targets during pretraining.

In the PredANN framework, a central assumption is that ANN representations that closely resemble human cortical representations can serve as effective teacher signals for EEG-based representation learning \cite{Akama2025}. Accordingly, we aimed to use ANN representations that are as neurophysiologically plausible as possible. A growing body of prior work has demonstrated that self-supervised learning (SSL) models capture human cortical representations more faithfully than supervised models, particularly in the auditory domain \cite{NEURIPS2022_d81ecfc8,vaidya2022selfsupervisedmodelsaudioeffectively,oota2023,oota2024deepneuralnetworksbrain}. These findings motivate the use of SSL-based models for extracting teacher representations from audio.

Moreover, prior studies have shown that the architectural properties of ANN models play a critical role in their correspondence with cortical representations. In particular, convolutional layers have been reported to exhibit lower predictive power for brain activity compared with Transformer layers \cite{NEURIPS2022_d81ecfc8}. In contrast, the functional hierarchy formed by Transformer layers has been shown to align well with the hierarchical organization of the human cerebral cortex \cite{NEURIPS2022_d81ecfc8}. Consistent with these observations, multiple studies suggest that Transformer-based models exhibit stronger representational similarity to cortical responses than alternative architectures \cite{NEURIPS2022_d81ecfc8,vaidya2022selfsupervisedmodelsaudioeffectively}.

Based on this converging evidence, we adopted Transformer-based SSL models to extract ANN representations from audio. 
Specifically, we used MuQ \cite{zhu2025muqselfsupervisedmusicrepresentation}, a masked-prediction-based Transformer model, to obtain acoustic representations, and MusicGen \cite{musicgen}, an autoregressive Transformer model, to derive expectation-related representations in the form of surprisal and entropy.
The detailed procedures for computing and discretizing these representations are described in the following sections and formalized in dedicated algorithms in \nameref{subsec:supp1} and \nameref{subsec:supp2}.

\subsection*{MuQ acoustic representation}

To construct ANN representations that primarily encode acoustic information, we employed MuQ \cite{zhu2025muqselfsupervisedmusicrepresentation}, a Transformer-based masked-prediction self-supervised music foundation model. MuQ has been shown to outperform representative SSL models such as MERT \cite{mert} and MusicFM \cite{musicfm} across nearly all tasks in the MARBLE benchmark \cite{marble}. Additionally, MuQ-MuLan, which incorporates MuQ as an audio encoder, achieves state-of-the-art performance in zero-shot music tagging, surpassing MuLan \cite{mulan2022} and Microsoft-CLAP 2023 \cite{elizalde2024clap}. These results indicate that MuQ provides a strong and expressive acoustic representation suitable as a teacher signal.

For each musical piece, continuous MuQ embeddings were extracted from 30-second audio chunks following the original MuQ preprocessing configuration. Audio signals were converted to mono by channel averaging, segmented into 30-second waveforms, and resampled to 24~kHz using a Kaiser-windowed sinc filter, implemented in the librosa package \cite{mcfee_librosa} via \texttt{librosa.resample} with \texttt{res\_type="kaiser\_fast"}. MuQ was then applied to each 30-second chunk, and the last-layer hidden representations were extracted, yielding a sequence of frame-wise embeddings at a temporal resolution of 25~Hz (one embedding every 40~ms). Formally, let $\mathcal{T}_{30\mathrm{s}} = \{1, \dots, 750\}$ denote the set of time frame indices within a 30-second audio chunk, where the temporal
resolution is 25~Hz (40-ms intervals). The resulting continuous acoustic representation $\mathbf{M}_{\mathrm{raw}}$ is defined as:
\[
\mathbf{M}_{\mathrm{raw}}
=
\bigl(\mathbf
{m}_{\mathrm{raw}_{t}}
\bigr)_{t \in \mathcal{T}_{30\mathrm{s}}},
\qquad
\mathbf
{m}_{\mathrm{raw}_{t}} \in \mathbb{R}^{1024},
\]
where each ${m}_{\mathrm{raw}_{t}}$ is a 1024-dimensional continuous acoustic feature vector at time frame $t$.

The use of 24~kHz sampling rate and fixed 30-second input length strictly follows the MuQ pretraining setup \cite{zhu2025muqselfsupervisedmusicrepresentation}. Extracting features under conditions that closely match the pretraining distribution reduces out-of-distribution effects and ensures that the resulting embeddings faithfully reflect the learned acoustic structure.

The continuous frame-wise acoustic representations $\mathbf{M}_{\mathrm{raw}}$, referred to as \emph{Raw MuQ}, are used as continuous inputs to the decoder during masked-prediction pretraining. In parallel, the same frame-wise embeddings are discretized to obtain a sequence of discrete acoustic tokens. Specifically, K-means clustering with $K=128$ is applied to the pooled set of all MuQ frame embeddings across all songs and chunks, yielding a 30-second discrete acoustic representation $\mathbf{M}_{\mathrm{disc}}$:
\[
\mathbf{M}_{\mathrm{disc}}
=
\bigl(
m_{\mathrm{disc}_{t}}
\bigr)_{t \in \mathcal{T}_{30\mathrm{s}}},
\qquad
m_{\mathrm{disc}_{t}} \in \{0,\dots,127\},
\]
where each $m_{\mathrm{disc}_{t}}$ is a scalar value representing the cluster index at time $t$. These \emph{Discretized MuQ} tokens constitute the 30-second discrete representation $\mathbf{M}_{\mathrm{disc}}$ and are later segmented into 3-second units $\mathbf{m}_{\mathrm{disc}}$ to serve as ground-truth targets for masked prediction.

Thus, the model explicitly distinguishes between continuous acoustic features $\mathbf{m}_{\mathrm{raw}_{t}}$ used as decoder inputs and discrete acoustic tokens $m_{\mathrm{disc}_{t}}$ used as supervision signals, while maintaining a one-to-one temporal correspondence between them. This frame-wise formulation provides the necessary foundation for constructing 3-second MuQ representations aligned with EEG segments in subsequent stages of the model. The detailed procedures for extracting continuous MuQ embeddings and performing K-means discretization are formalized in \nameref{subsec:supp1}.

\subsection*{MusicGen expectation-related representations}

To construct ANN representations that primarily encode expectation-related information, we employed MusicGen-large (facebook/musicgen-large) from the Audiocraft framework \cite{musicgen} as an autoregressive model.
MusicGen is the first model to achieve high-quality, long-duration, and controllable music generation using a single-stage Transformer language model. It is trained as a language model over discrete audio tokens produced by EnCodec \cite{encodec}, which applies residual vector quantization (RVQ) with multiple codebooks.

MusicGen operates on sequences of RVQ tokens obtained from EnCodec using four codebooks at a frame rate of 50~Hz (20~ms per step). Following the official MusicGen configuration, audio signals were first converted to mono by channel averaging and then resampled to 32~kHz using the function \texttt{audiocraft.data.audio\_utils.convert\_audio}.
The full waveform was then encoded into discrete RVQ token sequences with shape $(4, T)$, where $T$ denotes the number of frames in the song.

A key contribution of MusicGen lies in its treatment of multi-codebook RVQ tokens through codebook interleaving patterns. Copet et al.\ proposed and evaluated several patterns, including flattening, delay, parallel, and coarse first \cite{musicgen}. While the flattening pattern provides exact autoregressive factorization, the delay pattern uses an inexact but computationally efficient decomposition. Despite reducing computational cost by approximately a factor of four, evaluation showed that the delay pattern achieves similar generation quality to flattening. The MusicGen-large model used in this study is trained with the delay pattern, which we adopt throughout our experiments. We compute expectation features under an empty text condition, corresponding to the model’s unconditional distribution learned through classifier-free guidance training.

Let $z_t$ denote the token of the first RVQ codebook (k1) at time frame $t$, and let $p_\theta(v \mid C_t)$ denote the probability distribution over the k1 vocabulary predicted by the MusicGen language model, conditioned on the autoregressive context $C_t$. We define expectation-related quantities at the frame level as follows:
\[
s_{\mathrm{raw}_{t}} = -\log p_\theta(z_t \mid C_t),
\qquad
h_{\mathrm{raw}_{t}} = -\sum_{v} p_\theta(v \mid C_t)\log p_\theta(v \mid C_t).
\]
where $s_{\mathrm{raw}_{t}}$ and $h_{\mathrm{raw}_{t}}$ are scalar values representing the surprisal and entropy at time $t$, respectively.

In the default computation setting, these expectation-related quantities were computed by sliding a 3-second analysis window with a stride of 0.1~s over the full-length audio. Since the temporal resolution is 50~Hz, this 0.1-second stride exactly corresponds to a step size of 5 frames.
For a song with $T_{\mathrm{frames}}$ EnCodec frames, this yields
\[
N_{\mathrm{seg}} = \left\lfloor \frac{T_{\mathrm{frames}} - 150}{5} \right\rfloor + 1
\]
segments, where each segment consists of 150 frames corresponding to 3 seconds at 50~Hz. For segment index $j$, the segment boundaries are defined as $[s_j, e_j) = [5j, 5j + 150)$.

To compute expectation-related quantities for each segment, an autoregressive context window of length $W \in \{8,16,32\}$ seconds was constructed. In frame units, the window length is $W_f = 50W$, and the context spans $[e_j - W_f, e_j)$. When the context window extends before the beginning of the song, missing positions are padded with the MusicGen special token, which was explicitly learned during training to represent empty slots in the codebook interleaving schedule.

The context window tokens were fed into the MusicGen language model to obtain k1 logits with shape $(W_f, V)$, where $V$ is the k1 vocabulary size. The logits corresponding to the final 150 frames were extracted as the prediction targets. Using these logits, frame-wise surprisal and entropy values were computed for $t = 1,\dots,150$ within each segment yielding 3-second continuous expectation-related representations.
Formally, let $\mathcal{T}_{30\mathrm{s}} = \{1, \dots, 1500\}$ denote the set of frame indices within a 30-second chunk at 50~Hz. 
The full 30-second continuous expectation-related representations are defined as:
\[
\mathbf{S}_{\mathrm{raw}}
=
\bigl(
s_{\mathrm{raw}_{t}}
\bigr)_{t \in \mathcal{T}_{30\mathrm{s}}},
\qquad
\mathbf{H}_{\mathrm{raw}}
=
\bigl(
h_{\mathrm{raw}_{t}}
\bigr)_{t \in \mathcal{T}_{30\mathrm{s}}}.
\]
Let $\mathcal{T}_{3\mathrm{s}} = \{1, \dots, 150\}$ denote the set of frame indices within a 3-second window.
The 3-second expectation-related representations $\mathbf{s}_{\mathrm{raw}}$ and $\mathbf{h}_{\mathrm{raw}}$
used by the model are defined as contiguous subsequences of these 30-second representations:
\[
\mathbf{s}_{\mathrm{raw}}
=
\bigl(
s_{\mathrm{raw}_{t}}
\bigr)_{t \in \mathcal{T}_{3\mathrm{s}}},
\qquad
\mathbf{h}_{\mathrm{raw}}
=
\bigl(
h_{\mathrm{raw}_{t}}
\bigr)_{t \in \mathcal{T}_{3\mathrm{s}}}.
\]
The precise procedure for selecting and aligning these 3-second segments with the continuous EEG trials is detailed in the subsequent subsection (\hyperref[subsec:alignment]{Alignment between EEG and Music Features}).

Importantly, surprisal and entropy computation was restricted to the first codebook (k1). Under the delay-pattern schedule, higher codebooks are staggered such that a single processing step involves tokens from different time steps. 
Consequently, computing expectation-related quantities for higher codebooks (k2–k4) would involve conditioning on tokens from other codebooks that belong to future time steps (e.g., the context for k4 at time $t$ already contains k1 tokens from time $t+3$).
To avoid defining expectation-related features that implicitly reference such future information, we limit our computation to k1.
The validity and limitations of this approximation are discussed in the \hyperref[sec:discussion]{Discussion}.

The continuous frame-wise surprisal and entropy sequences, referred to as \emph{raw surprisal} and \emph{raw entropy}, are used as continuous decoder inputs during masked-prediction pretraining.
In parallel, these continuous values are discretized into 128 levels using equal-frequency (quantile) binning to obtain discrete expectation-related tokens used as ground-truth supervision. Specifically, for each feature type, all values across all songs and segments are pooled to form a one-dimensional set $u$. The global quantile edges are then computed as
\[
e_k = \mathrm{Quantile}(u, k/128), \quad k=0,\dots,128,
\]
where $e_k$ denotes the $k/128$ quantile of $u$. These values define $128$ quantile bins.  
Each continuous value $u$ is then assigned a bin index
$b \in \{0,\dots,127\}$ such that
\[
e_b \le u < e_{b+1},
\]
with the maximum value assigned to bin $127$.

By replacing each continuous value with its corresponding discrete token, the resulting discrete sequences (\emph{discretized surprisal} and \emph{discretized entropy}) maintain a strict one-to-one temporal correspondence with the continuous inputs.

The complete procedures for computing frame-wise surprisal and entropy and for quantile-based discretization are formalized in \nameref{subsec:supp2}.

\subsection*{Conservative chunk-based computation}

In addition to the default surprisal/entropy computation setting described above, we implemented a more conservative computation scheme in which all expectation-related quantities are computed independently within each 30-second audio chunk.
This alternative setting is not required for the validity of the main analyses, but serves as a supplementary control to examine the effect of strictly eliminating potential information leakage from acoustic context that is not available to the model at inference time.

Under the default setting, the EEG-based song ID classification model is trained and evaluated under the assumption that the musical piece itself is known during model training, and that short EEG segments (e.g., 3~s) are sampled from a longer, fixed musical context. In this scenario, it is not a realistic setting for one 30-second excerpt of a song to be treated as known while another 30-second excerpt from the same song is treated as entirely unknown. This assumption is analogous to the fact that Song ID labels are fixed and known during model training, and therefore using longer-range musical context during feature computation is not inherently problematic.

Nevertheless, we considered it informative to evaluate how model behavior and performance change when expectation-related features are computed under a stricter constraint that completely excludes acoustic context not explicitly associated with the EEG input used at inference time.
To this end, we adopted a conservative chunk-based computation scheme in which each song is first segmented into 30-second chunks using the same segmentation strategy applied to EEG data. For each 30-second chunk, k1 logits and continuous surprisal and entropy values are computed independently, without allowing autoregressive context to extend beyond the boundaries of the chunk. 

\begin{figure}[ht]
    \centering
    \includegraphics[width=0.85\linewidth]{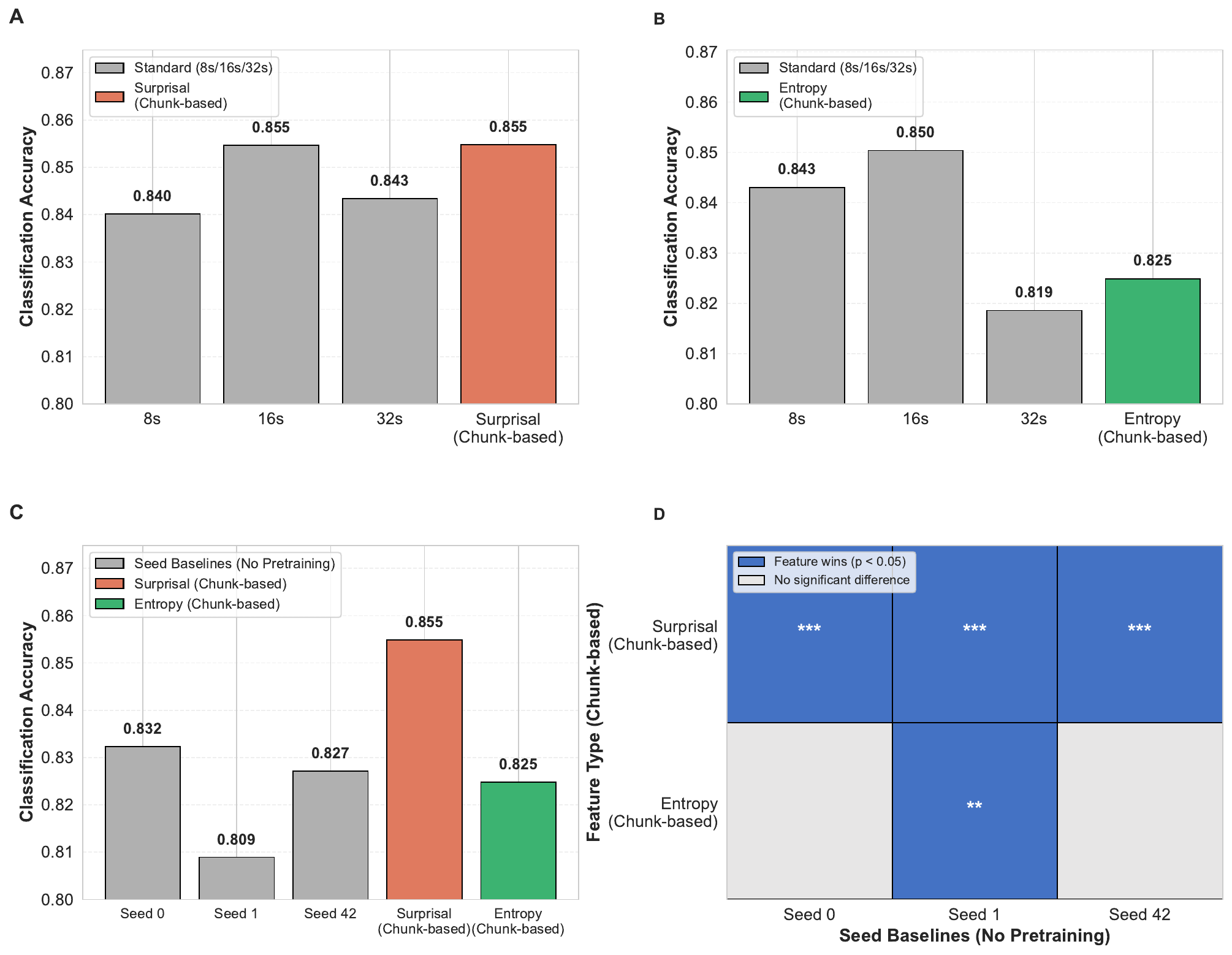}
    \caption{
        \textbf{Performance of the conservative chunk-based computation.}\\
        Panels A--D compare Surprisal and Entropy computed independently within 30-s audio chunks (see Methods, ``Conservative chunk-based computation'') with those obtained using standard context-length settings and seed baselines.\\ \textbf{A} Classification accuracy using Surprisal features. Grey bars denote the three standard computations (context lengths: 8~s, 16~s, 32~s); the orange bar indicates the conservative 30-s chunk-based computation. Chunk-based Surprisal yields accuracy comparable to the best-performing 16~s context. \textbf{B} Classification accuracy using Entropy features. Grey bars denote the three standard computations (context lengths: 8~s, 16~s, 32~s); the green bar indicates the conservative 30-s chunk-based computation. In contrast to Surprisal (panel A), chunk-based Entropy does not reach the performance of the 8~s or 16~s contexts, although it remains above the 32~s condition. \textbf{C} Comparison of three no-pretraining seed baselines (Seed 0/1/42, grey) with conservative Surprisal (orange) and Entropy (green). Surprisal exceeds all seed baselines, whereas Entropy surpasses Seed 1 but does not exceed Seed 0 or Seed 42. \textbf{D} McNemar's test heatmap comparing each seed column with each conservative feature row. Blue cells indicate feature superiority, light grey indicates no significant difference; asterisks indicate significance levels (** $p < 0.01$, *** $p < 0.001$). Conservative Surprisal shows significant improvements over all seed baselines, whereas conservative Entropy reaches significance for Seed 1 only, with no significant differences observed for Seed 0 or Seed 42.
    }
    \label{fig:supp1}
\end{figure}

Specifically, for each 30-second chunk, MusicGen \cite{musicgen} is applied to compute k1 logits at a temporal resolution of 20~ms, yielding 1500 frames per chunk. Surprisal and entropy are then computed frame-wise from these logits, and subsequently discretized into 128 bins using the same quantile-based discretization procedure as in the default setting. By construction, this approach prevents MusicGen from exploiting autoregressive context that spans across adjacent 30-second chunks, thereby strictly eliminating potential information leakage from unused acoustic context. The results of this conservative computation are presented in Figure~\ref{fig:supp1}.

\subsection*{Alignment between EEG and Music Features}
\phantomsection
\label{subsec:alignment}

Each 3-second EEG segment is aligned with music feature sequences covering the same temporal interval. EEG signals are sampled at $f_s = 125~\mathrm{Hz}$, and we denote the starting time of a 3-second EEG segment as $t_0$ (in seconds). Based on this starting time, corresponding segments of the music features are selected in a frame-wise manner, for both continuous (raw) and discretized representations.

For surprisal and entropy, which are computed at a temporal resolution of 20~ms (50~Hz), the starting frame index is defined as
\[
i_0 = \left\lfloor 50\, t_0 \right\rfloor,
\]
and a sequence of length 150 frames is extracted:
\[
\mathbf{s}_{\mathrm{raw}}
=
\bigl(
s_{\mathrm{raw}_{t}}
\bigr)_{t = i_0+1}^{i_0+150},
\qquad
\mathbf{h}_{\mathrm{raw}}
=
\bigl(
h_{\mathrm{raw}_{t}}
\bigr)_{t = i_0+1}^{i_0+150}.
\]
The corresponding discretized sequences are obtained using the same frame indices:
\[
\mathbf{s}_{\mathrm{disc}}
=
\bigl(
s_{\mathrm{disc}_{t}}
\bigr)_{t = i_0+1}^{i_0+150},
\qquad
\mathbf{h}_{\mathrm{disc}}
=
\bigl(
h_{\mathrm{disc}_{t}}
\bigr)_{t = i_0+1}^{i_0+150}.
\]
where $s_{\mathrm{disc}_{t}}, h_{\mathrm{disc}_{t}} \in \{0,\dots,127\}$. These discrete tokens are used as ground-truth supervision during pretraining.

For MuQ \cite{zhu2025muqselfsupervisedmusicrepresentation}, which provides acoustic embeddings at a temporal resolution of 40~ms (25~Hz), the starting frame index is defined as
\[
j_0 = \left\lfloor 25\, t_0 \right\rfloor,
\]
and a sequence of length 75 frames is extracted:
\[
\mathbf{m}_{\mathrm{raw}}
=
\bigl(
\mathbf{m}_{\mathrm{raw}_{t}}
\bigr)_{t = j_0+1}^{j_0+75}.
\]
The corresponding discretized MuQ sequence is obtained using the same temporal indices:
\[
\mathbf{m}_{\mathrm{disc}}
=
\bigl(
{m}_{\mathrm{disc}_{t}}
\bigr)_{t = j_0+1}^{j_0+75}.
\]
where $m_{\mathrm{disc}_{t}} \in \{0,\dots,127\}$.

In the default computation setting for surprisal and entropy, expectation features are stored as a collection of 3-second segments extracted with a stride of 0.1~s, and each segment is associated with metadata specifying its starting time $\texttt{segment\_start\_s}$. 
Given an EEG segment starting at $t_0$, we first identify all candidate surprisal/entropy segments that are fully contained within the same 30-second audio chunk. 
Among these candidates, we select the segment whose starting time is closest to $t_0$. The selected continuous sequences are used as decoder inputs during pretraining, while the corresponding discretized sequences are used as ground-truth labels on the decoder side.

This alignment procedure ensures that EEG segments and music features are temporally matched at the frame level while preserving a clear distinction between continuous inputs and discrete supervision signals. Implementation details are provided in the file \href{https://github.com/ShogoNoguchi/PredANNpp/blob/main/codes_3s/predann/datasets/preprocessing_eegmusic_dataset_3s.py}{\texttt{preprocessing\_eegmusic\_dataset\_3s.py}}.

\subsection*{Neural network architecture}

Recent progress in self-supervised pretraining has established masked modeling with Transformer architectures as a powerful paradigm for representation learning.
In natural language processing, BERT \cite{devlin-etal-2019-bert} demonstrated that predicting masked tokens enables the acquisition of rich bidirectional contextual representations, while the GPT series showed that autoregressive pretraining effectively captures long-range dependencies \cite{GPT, GPT2}.
In computer vision, this idea was reformulated as masked image modeling, where approaches such as MAE \cite{mae}, BEiT \cite{beit}, and SimMIM \cite{xie2021simmim} revealed that reconstructing missing content from partial observations leads to highly transferable representations.

A key step in transferring these ideas to EEG representation learning is exemplified by LaBraM \cite{jiang2024large}.
Inspired by masked modeling in vision, it reformulates EEG signals as collections of time--channel patches and applies Transformer-based masked reconstruction to large-scale EEG data.
Through architectural choices such as temporal patch encoders, learnable channel and time embeddings, and Pre-Norm self-attention, LaBraM demonstrated that masked modeling is an effective and scalable principle for EEG representation learning.

The original PredANN framework, by contrast, was built around a CLIP \cite{clip}-style contrastive learning objective using CNN-based encoders optimized with an InfoNCE loss.
While this design successfully aligned EEG representations with stimulus-derived ANN representations, it fundamentally differs from the generative and structural learning paradigm underlying recent masked autoencoding approaches.
Motivated by the success of LaBraM and masked modeling in other modalities, we therefore redesign PredANN into a Transformer-based masked modeling framework.

Specifically, we propose \textit{PredANN++}, which extends PredANN by replacing the CLIP-style CNN encoder with a masked autoencoding Transformer architecture.
Rather than discretizing EEG signals themselves as in LaBraM, PredANN++ treats discretized stimulus-derived representations (acoustic \cite{zhu2025muqselfsupervisedmusicrepresentation}, surprisal, and entropy \cite{musicgen,encodec}) as teacher sequences and formulates pretraining as masked prediction of these discrete targets.
This design preserves the original PredANN philosophy of learning EEG representations aligned with stimulus-level ANN representations, while adopting the masked generative learning paradigm shown to be effective in LaBraM.
The overall architecture is illustrated in Fig.~\ref{fig:architecture} and is formalized below.

\medskip

\textbf{EEG input representation.}
The EEG encoder takes as input a 3-second EEG segment
\[
\mathbf{X} \in \mathbb{R}^{128 \times 3 \times 125},
\]
where the dimensions correspond to channels, seconds, and samples per second, respectively.

\medskip

\textbf{Temporal patch embedding.}
Following the design principles of LaBraM, temporal structure is extracted before applying global self-attention.
The input EEG segment is processed by a temporal encoder consisting of three convolutional blocks, each followed by Group Normalization and GELU activation.
This encoder partitions the signal into channel--time patches and maps them into patch-level embeddings.
Formally, the temporal encoder produces
\[
\{\mathbf{e}_{c,s} \in \mathbb{R}^{128} \mid c = 1,\dots,128,\; s = 1,2,3\},
\]
resulting in $128 \times 3 = 384$ patch tokens.
Each patch token is then linearly projected into a shared embedding space of dimension $512$.

\medskip

\textbf{Channel and temporal embeddings.}
To encode spatial and temporal identity explicitly, we introduce learnable channel and second embeddings,
\[
\mathbf{E}^{\mathrm{ch}} \in \mathbb{R}^{128 \times 512}, \qquad
\mathbf{E}^{\mathrm{sec}} \in \mathbb{R}^{3 \times 512}.
\]
For each patch token, the corresponding channel and temporal embeddings are added.
A learnable \texttt{[CLS]} token is prepended to the sequence, yielding
\[
\mathbf{Z}_0 =
[\mathbf{z}_{\mathrm{cls}}, \mathbf{z}_1, \dots, \mathbf{z}_{384}]
\in \mathbb{R}^{385 \times 512}.
\]

\medskip

\textbf{EEG Transformer encoder.}
The token sequence is processed by a 2-layer Transformer encoder.
Each layer adopts a Pre-Norm formulation, following LaBraM, to stabilize optimization in shallow Transformer settings.
The self-attention operation is defined as
\[
\mathrm{Attention}(\mathbf{Q}, \mathbf{K}, \mathbf{V})
=
\mathrm{softmax}
\left(
\frac{\mathrm{LN}(\mathbf{Q})\,\mathrm{LN}(\mathbf{K})^{\top}}{\sqrt{d_{\mathrm{head}}}}
\right)
\mathbf{V},
\]
where $\mathrm{LN}(\cdot)$ denotes layer normalization and $d_{\mathrm{head}}$ is the per-head dimensionality.
The output corresponding to the \texttt{[CLS]} token is denoted as
\[
\mathbf{h}_{\mathrm{cls}} \in \mathbb{R}^{512}.
\]

\medskip

\textbf{Song ID classification branch.}
A lightweight supervised classification head is attached to $\mathbf{h}_{\mathrm{cls}}$.
Let
\[
\mathbf{z} = \mathrm{MLP}(\mathrm{LN}(\mathbf{h}_{\mathrm{cls}}))
\]
denote the classifier logits, where the MLP has a structure $512 \rightarrow 256 \rightarrow 10$, with Batch Normalization and ReLU activation in the hidden layer. 
The predicted label corresponds to the index of the largest logit value. This implementation corresponds to the module \texttt{projector1} (found in 
\href{https://github.com/ShogoNoguchi/PredANNpp/blob/main/codes_3s/predann/modules/EM_finetune.py}{\texttt{EM\_finetune.py}}, 
\href{https://github.com/ShogoNoguchi/PredANNpp/blob/main/codes_3s/predann/modules/Surprisal_multitask.py}{\texttt{Surprisal\_multitask.py}}, 
\href{https://github.com/ShogoNoguchi/PredANNpp/blob/main/codes_3s/predann/modules/Entropy_multitask.py}{\texttt{Entropy\_multitask.py}}, 
and 
\href{https://github.com/ShogoNoguchi/PredANNpp/blob/main/codes_3s/predann/modules/MuQ_multitask.py}{\texttt{MuQ\_multitask.py}}).
The classification loss is defined as
\[
\mathcal{L}_{C} = \mathrm{CE}(\mathbf{z}, y),
\]
where $y$ denotes the ground-truth Song ID and 
$\mathrm{CE}(\cdot)$ represents the cross-entropy applied to the logits.
This branch is introduced as an auxiliary objective during pretraining, following the design philosophy of SupMAE \cite{supmae}.

\medskip

\textbf{Masked teacher prediction branch.}
The main pretraining objective is to predict discretized music-derived teacher sequences under random masking.
Let
\[
\mathbf{m} = (m_1,\dots,m_{N_M}), \qquad m_i \in \{0,\dots,127\},
\]
denote the discrete teacher sequence aligned to the EEG segment, where $N_M = 150$ for Surprisal/Entropy and $N_M = 75$ for MuQ.

For each position $i$, the decoder input is defined as
\[
\mathbf{u}_i =
\begin{cases}
\mathbf{e}_{\mathrm{mask}}, & i \in \mathcal{M}, \\
\mathbf{W}\mathbf{z}_i, & i \notin \mathcal{M},
\end{cases}
\]
where $\mathcal{M}$ is a randomly sampled mask set covering $50\%$ of positions,
$\mathbf{e}_{\mathrm{mask}} \in \mathbb{R}^{512}$ is a learnable mask embedding, and
$\mathbf{z}_i$ is the corresponding continuous teacher feature
(MuQ: $\mathbf{z}_i \in \mathbb{R}^{1024}$; Surprisal/Entropy: $\mathbf{z}_i \in \mathbb{R}$),
linearly projected to $512$ dimensions by $\mathbf{W}$.
Learnable temporal positional embeddings are added to $\{\mathbf{u}_i\}$.

By randomly varying the masked regions in each batch, the model is compelled to capture the global relational structure from a partially observed context. This approach serves as a powerful regularizer, which is essential for preventing overfitting when training on relatively small EEG datasets.

The decoder input is formed by concatenating encoder tokens and teacher tokens, and processed by a 2-layer Transformer decoder.
For each teacher position $i$, the decoder outputs logits
\[
\hat{\mathbf{m}}_i \in \mathbb{R}^{128}.
\]
The masked teacher prediction loss is defined as
\[
\mathcal{L}_{M}
=
\frac{1}{|\mathcal{M}|}
\sum_{i \in \mathcal{M}}
\mathrm{CE}(\hat{\mathbf{m}}_i, m_i).
\]

\medskip

\textbf{Discretisation and learning stability.}
Prior to pretraining, we transform the continuous teacher representations into discrete tokens to ensure stable learning signals. Acoustic (MuQ) features are converted via $k$-means clustering ($k=128$), while surprisal and entropy are partitioned into 128 equally-populated bins via quantile-based binning. This discretization acts as an implicit denoising mechanism, discouraging the decoder from fitting low-level noise in raw-valued features. Predicting discrete tokens via cross-entropy provides more stable optimization than regressing continuous values.

\medskip

\textbf{Multitask pretraining objective.}
The overall pretraining objective is defined as
\[
\mathcal{L}
=
1.0 \cdot \mathcal{L}_{C}
+
0.1 \cdot \mathcal{L}_{M}.
\]
This formulation follows the core idea of SupMAE: combining a masked generative objective with a supervised classification branch \cite{supmae}.
In PredANN++, masked teacher prediction enforces local, time-resolved alignment between EEG and stimulus-derived representations, while the classification objective promotes global semantic organization of EEG representations.

\medskip

\textbf{Relation to prior works.}
Compared with PredANN, which is built upon a contrastive CNN architecture optimized via an InfoNCE objective, PredANN++ adopts a Transformer-based generative masked modeling paradigm.
This change represents a shift from contrastive alignment toward structured generative pretraining, following the design principles that have proven effective in masked autoencoding frameworks across multiple modalities \cite{devlin-etal-2019-bert, mae, beit, mae_audio1, jiang2024large}.

In comparison with LaBraM, our model employs a substantially shallower Transformer encoder consisting of only two layers, while increasing the embedding dimension to 512.
This configuration is chosen to mitigate underfitting in shallow architectures while retaining sufficient representational capacity.
Furthermore, whereas LaBraM discretizes EEG signals themselves and performs masked reconstruction in the EEG domain, PredANN++ operates on discretized stimulus-derived teacher representations.
This design choice builds upon the vector-quantized generative modeling framework already established in LaBraM and reuses discrete latent variables as efficient generative targets.

\subsection*{Training schedule and optimization}

Multitask pretraining was conducted for 10{,}000 epochs under a unified experimental setting.
All multitask models were trained with a fixed random seed of 42 to ensure reproducibility.
Optimization was performed using Adam \cite{adam2017} for both pretraining and fine-tuning stages, with a learning rate of 0.003 and a batch size of 48.
The checkpoint obtained at the final pretraining epoch was used as the initialization for subsequent fine-tuning.
Across all teacher types (Acoustic, Surprisal, and Entropy), the EEG encoder architecture was kept identical, ensuring that differences in downstream performance arise solely from the choice of teacher representation rather than architectural variation.

After multitask pretraining, the model was fine-tuned using only the Song ID classification objective.
Specifically, encoder weights were initialized from the multitask pre-trained checkpoint, while all decoder-related parameters—including the mask token embeddings, temporal positional embeddings, teacher projection layers, and Transformer decoder blocks—were discarded.
The remaining encoder and Song ID classification head were then trained for an additional 3{,}500 epochs, resulting in a Song ID classifier specific to each teacher representation.

As a baseline, we trained the same encoder-only architecture for Song ID classification from scratch without multitask pretraining (\emph{Fullscratch}).
For a fair comparison, the Fullscratch models were trained for the same number of epochs as the fine-tuning stage, namely 3{,}500 epochs, using identical optimization settings.
To quantify the effect of initialization diversity, Fullscratch models were trained with multiple random seeds (e.g., 0, 1, and 42).
These independently trained models were further used to evaluate seed-based ensemble effects, providing a direct comparison with ensembles constructed from models pretrained with different teacher representations.

\subsection*{Ensemble inference}

At inference time, we evaluate the complementarity of learned EEG representations using an equal-weight Deep Ensemble strategy ~\cite{DeepEnsemble2017}.
Let $K$ denote the number of independently trained classifiers included in the ensemble, with $K \in \{2,3\}$.
Given an EEG input $\mathbf{x}$, each classifier $f^{(k)}$ outputs a vector of logits
\[
\hat{\mathbf{y}}^{(k)}(\mathbf{x}) \in \mathbb{R}^{10}, \qquad k = 1,\dots,K.
\]

For each classifier, logits are converted into class probabilities via the softmax function,
\[
p^{(k)}(y \mid \mathbf{x}) = \mathrm{softmax}\!\left(\hat{\mathbf{y}}^{(k)}(\mathbf{x})\right)_y.
\]
The ensemble predictive distribution is then obtained by uniformly averaging the class probabilities across classifiers,
\[
p_{\mathrm{ens}}(y \mid \mathbf{x})
=
\frac{1}{K} \sum_{k=1}^{K} p^{(k)}(y \mid \mathbf{x}).
\]
The final ensemble prediction is given by
\[
\hat{y}_{\mathrm{ens}} = \arg\max_{y} \, p_{\mathrm{ens}}(y \mid \mathbf{x}).
\]

This ensemble rule is applied to two distinct model groups.
First, we construct ensembles from models pretrained with different teacher representations (Acoustic, Surprisal, and Entropy), thereby probing the complementarity of stimulus-aligned EEG representations learned along different information axes.
Second, we apply the same ensemble strategy to \emph{Fullscratch} models trained from random initialization with different random seeds.
By comparing ensembles formed across teacher representations with those formed across random seeds, we isolate gains arising from representational complementarity rather than mere initialization diversity.

To clearly characterize the effect of ensemble size, we primarily report results for $K=2$ and $K=3$, which allow direct assessment of how performance evolves as additional models are integrated.

\subsection*{Evaluation protocol}

All evaluation results are reported under a fixed data split with \texttt{split\_seed = 42}.
The dataset is partitioned in a song-stratified manner, and performance is evaluated on the resulting validation set.
This design ensures that all reported accuracies are directly comparable across models and training conditions.

For evaluation, the EEG data loader is configured deterministically.
Within each 8-second sliding window, the 3-second EEG segment is always extracted from the temporal center of the window.
As a result, repeated evaluations yield identical EEG segments and perfectly aligned teacher segments, eliminating stochastic variation during inference.

The primary evaluation metric is Song ID classification accuracy.
To ensure reproducibility of ensemble inference and statistical testing, the evaluation code caches the per-sample logits produced by each model.
These cached logits are reused for constructing ensembles and performing statistical comparisons, guaranteeing that all analyses are conducted on exactly the same set of predictions.

To assess whether the difference in accuracy between two models (or between a single model and an ensemble) evaluated on the same samples is statistically significant, we employ McNemar's test with an exact, two-sided formulation.
Given two predictors A and B evaluated on the same $N$ samples, we construct a $2 \times 2$ contingency table consisting of:
$a$, the number of samples correctly classified by both A and B;
$b$, the number of samples correctly classified by A but misclassified by B;
$c$, the number of samples correctly classified by B but misclassified by A; and
$d$, the number of samples misclassified by both A and B.
The test is performed exactly on the discordant pair $(b, c)$, and statistical significance is determined at the threshold $p < 0.05$.

Throughout this study, claims of ``statistically significant improvement'' are based exclusively on the exact $p$-values obtained from McNemar's test.
Implementation details of the evaluation procedure, including ensemble construction and statistical testing, are provided in \href{https://github.com/ShogoNoguchi/PredANNpp/blob/main/codes_3s/analysis/evaluate.py}{\texttt{evaluate.py}}.

\subsection*{Implementation details}

All models were implemented using PyTorch and PyTorch Lightning.
The released code has been validated under the following environment:
Ubuntu~20.04, Python~3.8, PyTorch~2.1.2, and PyTorch Lightning~1.4.0,
with CUDA runtime~11.8 (PyTorch cu118 build).
All experiments were conducted on an NVIDIA RTX~A6000 GPU with 48~GB of VRAM.
\section*{Data availability}
The Naturalistic Music EEG Dataset – Tempo (NMED-T) \cite{losorelli2017nmed} analyzed in this study is publicly available at: \url{https://exhibits.stanford.edu/data/catalog/jn859kj8079}.

\section*{Code availability}
The complete source code for dataset preprocessing, model training, evaluation, and implementation of the proposed method is publicly available at \url{https://github.com/ShogoNoguchi/PredANNpp}. The repository provides all scripts necessary to reproduce the experiments, including multitask pretraining, finetuning, evaluation pipelines, and a Gradio-based inference demo. Furthermore, we provide a comprehensive project page at \url{https://shogonoguchi.github.io/PredANNpp/}, which serves as an online supplement providing the project overview, results tables, pretrained checkpoints, interactive audio-synchronized feature visualizations, Gradio demo videos, and dataset information.
The code is released under the CC-BY-SA 4.0 license.

\section*{Funding}
This work was supported by Sony Computer Science Laboratories, Inc.

\bibliography{sn-bibliography}

\section*{Supplementary information}
\phantomsection
\label{sec:supp}

\subsection*{Supplementary Note 1: Algorithmic details of MuQ extraction}
\label{subsec:supp1}

In Algorithm~\ref{algorithm:muq-continuous}, the input waveform is transformed explicitly as
$\mathbf{x} \rightarrow \mathbf{x}_1 \rightarrow \mathbf{x}_2 \rightarrow \mathbf{x}_3$.
In Algorithm~\ref{algorithm:muq-discretize}, K-means is run with K-means++ initialization,
\texttt{random\_state}=0, and \texttt{n\_init}=10.

\begin{algorithm}[H]
\caption{Extraction of frame-wise continuous MuQ embeddings from 30-second audio}
\label{algorithm:muq-continuous}
\begin{algorithmic}[1]
\Require Audio waveform $\mathbf{x}$ and pre-trained MuQ model with frozen parameters $\phi$
\Ensure Frame-wise continuous MuQ representation $\mathbf{M}_{\mathrm{raw}} = \bigl(\mathbf{m}_{\mathrm{raw}_{t}}\bigr)_{t=1}^{750}$, where $\mathbf{m}_{\mathrm{raw}_{t}} \in \mathbb{R}^{1024}$
\State Convert $\mathbf{x}$ to mono and resample to $24~\mathrm{kHz}$ to get $\mathbf{x}_1$
\State Trim or zero-pad $\mathbf{x}_1$ to exactly $30$ seconds to get $\mathbf{x}_2$ \Comment{MuQ expects fixed-length 30-s inputs}
\State Normalize $\mathbf{x}_2$ according to MuQ preprocessing to get $\mathbf{x}_3$
\State Obtain MuQ hidden states $\mathbf{H}$ from $\mathrm{MuQ}_{\phi}(\mathbf{x}_3)$
\State Extract the final-layer representation $\mathbf{H}_{\mathrm{last}} \in \mathbb{R}^{1 \times 750 \times 1024}$ from $\mathbf{H}$
\For{$t \gets 1$ to $750$}
    \State $\mathbf{m}_{\mathrm{raw}_{t}} \gets \mathbf{H}_{\mathrm{last},1,t,:}$
\EndFor
\State \Return $\mathbf{M}_{\mathrm{raw}}$
\end{algorithmic}
\end{algorithm}

\begin{algorithm}[H]
\caption{K-means discretization of frame-wise MuQ embeddings}
\label{algorithm:muq-discretize}
\begin{algorithmic}[1]
\Require Pooled MuQ frame embeddings $\mathcal{E} = \bigcup_{\text{songs}} \bigcup_{\text{chunks}} \{\mathbf{m}_{\mathrm{raw}_{t}}\}$ and number of clusters $K = 128$
\Ensure For each 30-second chunk $c \in \{1,\dots,C\}$, a discrete MuQ sequence $\mathbf{M}^{(c)}_{\mathrm{disc}} = \bigl(m^{(c)}_{\mathrm{disc}_{t}}\bigr)_{t=1}^{750}$, where $m^{(c)}_{\mathrm{disc}_{t}} \in \{0,\dots,K-1\}$
\State Fit K-means to $\mathcal{E}$ with $K$ clusters
\State Obtain cluster centroids $\{\mathbf{c}_k\}_{k=0}^{K-1}$
\For{each 30-second chunk $c$}
    \For{$t \gets 1$ to $750$}
        \State $m^{(c)}_{\mathrm{disc}_{t}} \gets \argmin_{k \in \{0,\dots,K-1\}} \lVert \mathbf{m}^{(c)}_{\mathrm{raw}_{t}} - \mathbf{c}_k \rVert_2$
    \EndFor
\EndFor
\State \Return $\{\mathbf{M}^{(c)}_{\mathrm{disc}}\}_{c=1}^{C}$
\end{algorithmic}
\end{algorithm}

\subsection*{Supplementary Note 2: Algorithmic details of computing surprisal, entropy, and discretization}
\label{subsec:supp2}

For Algorithm~\ref{algorithm:surp_ent_cont}, we write
$\mathbf{s}^{(j)}_{\mathrm{raw}} = \bigl(s^{(j)}_{\mathrm{raw},t}\bigr)_{t=1}^{150}$
and
$\mathbf{h}^{(j)}_{\mathrm{raw}} = \bigl(h^{(j)}_{\mathrm{raw},t}\bigr)_{t=1}^{150}$
for the 3-second surprisal and entropy sequences of segment $j$.
For Algorithm~\ref{algorithm:surp_ent_disc}, we write
$\mathbf{s}^{(j)}_{\mathrm{disc}} = \bigl(s^{(j)}_{\mathrm{disc},t}\bigr)_{t=1}^{150}$
and
$\mathbf{h}^{(j)}_{\mathrm{disc}} = \bigl(h^{(j)}_{\mathrm{disc},t}\bigr)_{t=1}^{150}$
for the corresponding discretized sequences.
Here, $\BinIndex(u;E)$ returns the unique bin index $b \in \{0,\dots,B-1\}$ such that
$E_b \le u < E_{b+1}$, with the maximum value assigned to bin $B-1$.

\begin{algorithm}[H]
\caption{Sliding-window computation of frame-wise surprisal and entropy using MusicGen k1}
\label{algorithm:surp_ent_cont}
\begin{algorithmic}[1]
\Require Full-length audio waveform $\mathbf{x}$ and context window length $W \in \{8,16,32\}$ seconds
\Ensure Segment-wise continuous predictive representations $\mathbf{S}_{\mathrm{raw}} = \bigl\{\mathbf{s}^{(j)}_{\mathrm{raw}}\bigr\}_{j=0}^{N_{\mathrm{seg}}-1}$ and $\mathbf{H}_{\mathrm{raw}} = \bigl\{\mathbf{h}^{(j)}_{\mathrm{raw}}\bigr\}_{j=0}^{N_{\mathrm{seg}}-1}$
\State Convert $\mathbf{x}$ to mono and resample to $32~\mathrm{kHz}$ to get $\mathbf{x}_1$
\State Encode $\mathbf{x}_1$ using EnCodec to obtain $\mathbf{z} \in \mathbb{N}^{4 \times T_{\mathrm{frames}}}$ \Comment{4 codebooks at 50~Hz}
\State $L \gets 150$ \Comment{3-s segment length in frames}
\State $d \gets 5$ \Comment{0.1-s stride in frames}
\State $N_{\mathrm{seg}} \gets \left\lfloor \frac{T_{\mathrm{frames}} - L}{d} \right\rfloor + 1$
\For{$j \gets 0$ to $N_{\mathrm{seg}}-1$}
\State $s_j \gets j d$
\State $e_j \gets s_j + L$
\State $W_f \gets 50 W$
    \State Construct the context token sequence from frames $[e_j - W_f,\; e_j)$ to get $\mathbf{z}^{(j)}_{\mathrm{ctx}}$
    \If{$e_j - W_f < 0$}
        \State Left-pad the missing prefix of $\mathbf{z}^{(j)}_{\mathrm{ctx}}$ with the MusicGen special token
    \EndIf
    \State Compute k1 logits from MusicGen conditioned on $\mathbf{z}^{(j)}_{\mathrm{ctx}}$ under the empty-text condition
    \State Extract the k1 logits corresponding to the final $L$ frames
    \For{$t \gets 1$ to $L$}
        \State $s^{(j)}_{\mathrm{raw},t} \gets -\log p_{\theta}(z_{s_j+t} \mid C_{j,t})$
        \State $h^{(j)}_{\mathrm{raw},t} \gets -\sum_v p_{\theta}(v \mid C_{j,t})\log p_{\theta}(v \mid C_{j,t})$
    \EndFor
\EndFor
\State \Return $\mathbf{S}_{\mathrm{raw}}$, $\mathbf{H}_{\mathrm{raw}}$
\end{algorithmic}
\end{algorithm}

\begin{algorithm}[H]
\caption{Quantile-based discretization of surprisal and entropy}
\label{algorithm:surp_ent_disc}
\begin{algorithmic}[1]
\Require Segment-wise continuous representations $\mathbf{S}_{\mathrm{raw}}$ and $\mathbf{H}_{\mathrm{raw}}$, and number of bins $B = 128$
\Ensure Segment-wise discrete representations $\mathbf{S}_{\mathrm{disc}} = \bigl\{\mathbf{s}^{(j)}_{\mathrm{disc}}\bigr\}_{j=0}^{N_{\mathrm{seg}}-1}$ and $\mathbf{H}_{\mathrm{disc}} = \bigl\{\mathbf{h}^{(j)}_{\mathrm{disc}}\bigr\}_{j=0}^{N_{\mathrm{seg}}-1}$
\State Pool all Surprisal values across all segments into $\mathcal{S}$
\State Pool all Entropy values across all segments into $\mathcal{H}$
\For{$k \gets 0$ to $B$}
    \State $q_k \gets k / B$
    \State $E^{(S)}_k \gets \Quantile(\mathcal{S}, q_k)$
    \State $E^{(H)}_k \gets \Quantile(\mathcal{H}, q_k)$
\EndFor
\For{each segment $j$}
    \For{$t \gets 1$ to $150$}
        \State $s^{(j)}_{\mathrm{disc},t} \gets \BinIndex\!\left(s^{(j)}_{\mathrm{raw},t}; E^{(S)}\right)$
        \State $h^{(j)}_{\mathrm{disc},t} \gets \BinIndex\!\left(h^{(j)}_{\mathrm{raw},t}; E^{(H)}\right)$
    \EndFor
\EndFor
\State \Return $\mathbf{S}_{\mathrm{disc}}$, $\mathbf{H}_{\mathrm{disc}}$
\end{algorithmic}
\end{algorithm}

\subsection*{Supplementary Note 3: Qualitative inspection of context-dependent surprisal/entropy time courses}
\label{subsec:supp3}

\begin{figure}[p]
    \centering
    \includegraphics[height=0.82\textheight]{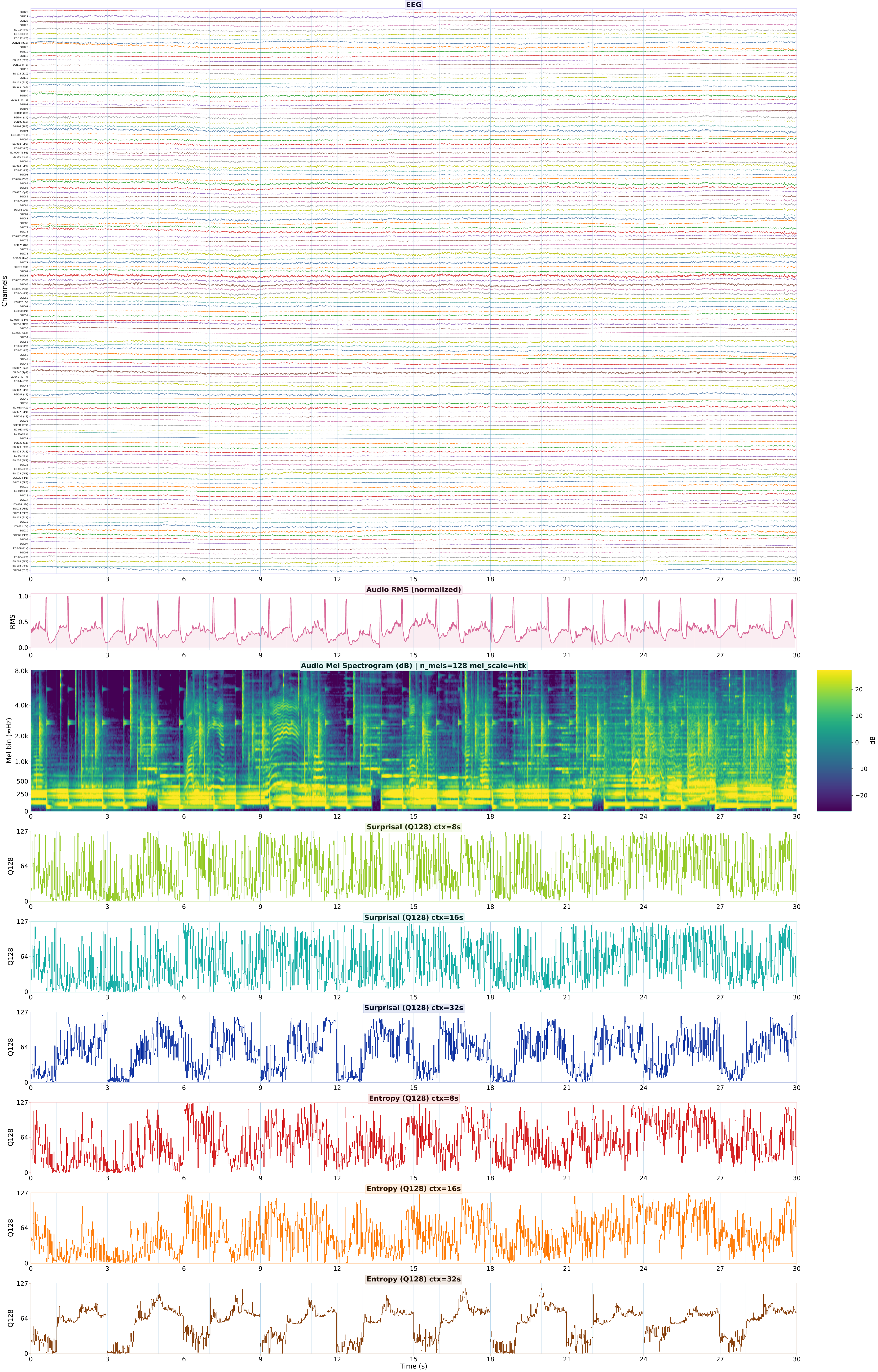}
    \caption{
        \textbf{Example multi-panel visualization for Song 21 (30--60~s), Subject 2.}\\
        Panels show (top to bottom) scalp EEG from Subject~2 recorded with the EGI 128-channel HydroCel Geodesic Sensor Net (all channels stacked), the normalized short-time RMS envelope, the mel-spectrogram in decibels (128 mel bins; HTK mel scale), and MusicGen-k1-derived surprisal and entropy trajectories discretized into Q128 bins for context windows of 8, 16, and 32 seconds (see \nameref{subsec:supp3} for qualitative inspection).
    }
    \label{fig:supp3}
\end{figure}

Figure~\ref{fig:supp3} provides an example visualization in which MusicGen-derived surprisal and entropy are displayed together with acoustic representations and human EEG from Subject~2 on a common time axis.

In this visualization, scalp EEG signals recorded with the EGI 128-channel HydroCel Geodesic Sensor Net are displayed using stacked traces for all available channels. This representation allows the temporal dynamics of large-scale scalp EEG activity to be inspected together with the acoustic and model-derived features on the same time axis.

A consistent qualitative trend across context lengths can be observed: when the context window increases from 8~s to 16~s and 32~s, the surprisal and entropy trajectories appear progressively smoother and vary on slower time scales. This visual smoothing is compatible with the idea that longer contexts integrate information over longer musical histories, and therefore the resulting predictive quantities may show reduced sensitivity to short-term fluctuations.

The surprisal and entropy trajectories can also be visually compared with the acoustic representations (RMS envelope and mel-spectrogram) shown above, allowing qualitative inspection of how the predictive quantities evolve relative to the stimulus structure.

For clarity, the mel-spectrogram panel titled ``Audio Mel Spectrogram (dB) | n\_mels=128 mel\_scale=htk'' is computed by applying a 128-bin mel filterbank (HTK mel scale) to the power spectrogram and converting the resulting mel-band power to decibels using $10\log_{10}(\cdot)$ for visualization. The RMS panel is computed as a short-time root-mean-square envelope of the waveform and is min--max normalized to the range $[0,1]$ within the plotted 30-second window.

To further facilitate qualitative inspection of how predictive quantities relate to the acoustic structure of the music, we provide an interactive web-based visualization at \url{https://shogonoguchi.github.io/PredANNpp/#syncviz}. 
This interface enables synchronized playback of the audio signal together with the corresponding MusicGen-derived surprisal and entropy trajectories, with the RMS envelope optionally displayed.
The audio examples used in this visualization are drawn from the MTG-Jamendo dataset.

\clearpage
\section*{Supplementary figures}


\clearpage

\clearpage
\section*{Author information}

\subsection*{Authors and Affiliations}

\textbf{Sony Computer Science Laboratories, Inc., Tokyo, Japan}\\
Shogo Noguchi, Taketo Akama, Tai Nakamura, Shun Minamikawa \& Natalia Polouliakh

\subsection*{Contributions}

S.N. and T.A. jointly conceptualized the study. T.A. proposed the initial framework and research direction. S.N. refined and operationalized the methodology, designed the model architecture and experimental protocol, and led the software implementation. S.N. and T.N. conducted the experiments with feedback from T.A. ; T.N. additionally contributed to visualization and statistical testing, and assisted with supporting code for experiments and analyses. S.M. helped organize and finalize the codebase. S.N. wrote the first draft of the manuscript with contributions from T.N. and T.A.; S.N. and T.N. created the figures and tables, with feedback from T.A. T.A. and N.P. reviewed and edited the manuscript. N.P. advised the research and organized the research project.

\subsection*{Corresponding author}

Correspondence to Shogo Noguchi and Taketo Akama.

\subsection*{Competing interests}

The authors declare no competing interests.

\end{document}